\documentclass{article}

\usepackage[final,main]{neurips_2026}
\usepackage[utf8]{inputenc}
\usepackage[T1]{fontenc}
\usepackage{courier}
\usepackage{hyperref}
\usepackage{url}
\usepackage{booktabs}
\usepackage{amsmath}
\usepackage{amssymb}
\usepackage{amsfonts}
\usepackage{amsthm}
\usepackage{aliascnt}
\usepackage{mathtools}
\usepackage{nicefrac}
\usepackage{microtype}
\usepackage{xcolor}
\usepackage{graphicx}
\usepackage{array}
\usepackage{longtable}
\usepackage{multirow}
\usepackage{algorithm}
\usepackage{algpseudocode}
\usepackage{float}
\usepackage{subcaption}
\usepackage{placeins}
\usepackage{caption}
\usepackage{float}
\usepackage[capitalize,noabbrev]{cleveref}

\theoremstyle{plain}

\newaliascnt{proposition}{theorem}
\newtheorem{proposition}[proposition]{Proposition}
\aliascntresetthe{proposition}

\theoremstyle{definition}

\theoremstyle{remark}

\crefname{assumption}{Assumption}{Assumptions}
\Crefname{assumption}{Assumption}{Assumptions}
\crefname{proposition}{Proposition}{Propositions}
\Crefname{proposition}{Proposition}{Propositions}

\newcommand{\R}{\mathbb{R}}

\newcommand{\F}{\mathcal{F}}
\DeclareMathOperator*{\argmin}{arg\,min}

\DeclareMathOperator{\diag}{diag}

\newcommand{\Frob}[1]{\left\| #1 \right\|_{F}}

\newcommand{\Delt}{\Delta}
\newcommand{\Wmat}{W}
\newcommand{\Vbasis}{V}
\newcommand{\overlapfn}{\mathrm{overlap}}
\newcommand{\rootscore}{\mathrm{root\_score}}

\newcommand{\Nw}{N_w}

\title{Witness Overlap: Directional Provenance Inside Open-Weight Model Families}

\author{
Siyuan Li\textsuperscript{1},
Haoxuan Zeng\textsuperscript{1},
Xin Luo\textsuperscript{2},
Fernando Jia\textsuperscript{1},
Florence Li\textsuperscript{1},
\\
\bfseries
Zhengyang Geng\textsuperscript{3},
Zico Kolter\textsuperscript{3},
Tai Sing Lee\textsuperscript{3},
Tianqin Li\textsuperscript{3}
\\[0.5em]
\normalfont
\textsuperscript{1}ModelOS  \textsuperscript{2}University of Michigan \\
\textsuperscript{3}Carnegie Mellon University
\\
}

\begin{document}

\maketitle

\begin{abstract}

Open-weight models are often released, fine-tuned, aligned, merged, and re-released, making provenance audits ask not only whether checkpoints are related, but also which checkpoint came first. Many existing model-provenance methods are designed for a base-known audit setting: given a victim or source model, they test whether a suspect model is related to it. Although these audits are framed as source-to-suspect tests, their underlying evidence is often symmetric, relying on representation similarity, weight similarity, behavioral fingerprints, or correlation statistics. Symmetric pairwise comparisons can detect relatedness, but they cannot by themselves orient relationship between checkpoints A and B. 
We therefore introduce a local geometric comparison: instead of comparing two checkpoints directly, we add a third same-family checkpoint as a witness and compare the geometry around each candidate endpoint. Direction is inferred by asking which candidate behaves more like a branching parent. Motivated by this idea, and by the empirically observed asymmetry between parent-anchored and child-anchored witness-overlap distributions, we propose Witness Overlap, a prompt-free, training-free white-box test for directional provenance. On 176 LLM checkpoints from 16 families, our one-witness test orients 95.3\% of parent–child decisions using Frobenius cosine. We further evaluate root identification, sibling discrimination, generalizations to VLM and diffusion families, and chain-structured ordering. The signal is robust to weight noise and sparse pruning, with a proposed SVD weight reduction variant showing greater robustness than Frobenius cosine.
\end{abstract}

\section{Introduction}
\label{sec:intro}
Open-weight large language models are being released, fine-tuned, and aligned at an accelerating pace, making provenance important for audits, license disputes, and safety investigations. Most existing work frames provenance as a fixed-source problem: given a known base model, determine whether a suspect checkpoint was derived from it. Many white-box, black-box, and behavioral methods are effective for this task by comparing weights, representations, activations, outputs, prompts, likelihoods, or generated-text statistics~\citep{zhang2025reef,zhu2025independence,liu2026fnf,zhang2026attndiff,nikolic2025mptllm,hu2025llmprint,kuditipudi2025palimpsestic}.

However, the fixed-known-base assumption becomes limiting as model hubs contain many checkpoints from the same family. An auditor may not only ask \emph{whether two checkpoints are related}, but also \emph{which checkpoint came first}. Most pairwise provenance scores, such as weight distance, representation similarity, and related measures, are symmetric under swapping $A$ and $B$, so they can detect relatedness but cannot orient a parent--child edge. 

\begin{figure}[t]
    \centering
    \includegraphics[width=\textwidth]{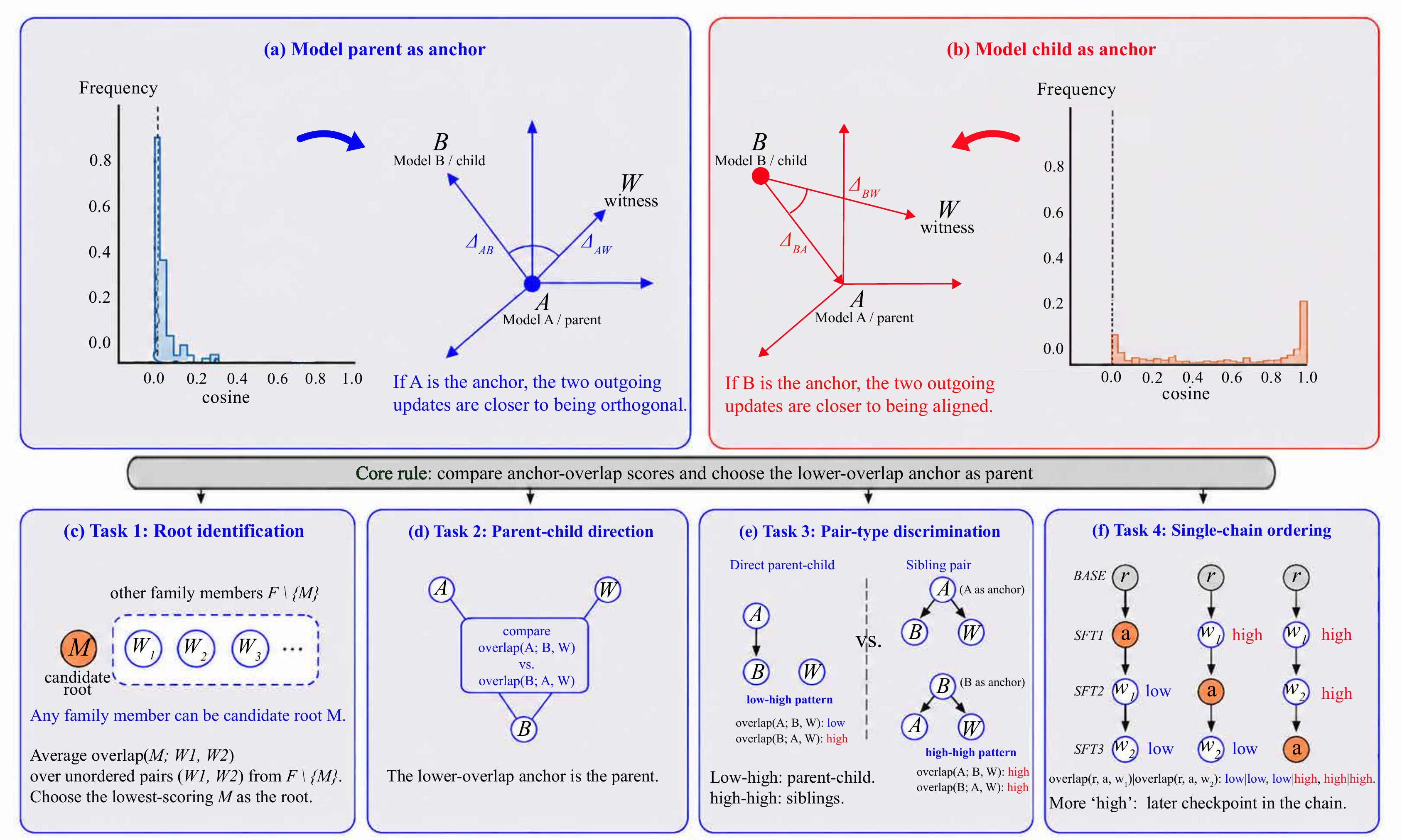} \caption{\textbf{Anchor-Conditioned Witness Overlap Across Four Provenance Tasks.}
    \textsc{Witness Overlap} scores each anchor-target-witness triplet by comparing the two update directions from the anchor.
\textbf{(a)} When the parent is used as the anchor, the left frequency histogram shows that cosine overlaps between updates to different descendants concentrate near low values; the right schematic illustrates these weakly aligned outgoing directions. The lower-overlap endpoint is predicted as the parent.
\textbf{(b)} When a child is used as the anchor, the overlap distribution shifts higher because the directions to the parent and to a sibling share a reverse component toward the parent; the left schematic illustrates this aligned geometry. The higher-overlap endpoint is predicted as the child.
\textbf{(c)--(f)} The same anchor-conditioned low/high pattern supports root identification, parent-child orientation, pair-type discrimination, and single-chain ordering: select the lowest-scoring candidate root, compare endpoint anchor scores, separate low-high from high-high patterns, and sort checkpoints by average overlap relative to a known base to identify the chain.}
    \label{fig:hero_witness_overlap}
\end{figure}

Our method is motivated by a simple weight-space observation. Neural network checkpoints lie in a high-dimensional parameter space, where different fine-tuning runs from the same parent can move along distinct directions~\citep{yadav2023ties}. As a result, updates from a shared parent to independently fine-tuned children are often weakly aligned, with low cosine similarity and large angles (\Cref{fig:hero_witness_overlap}a). However, when the comparison is anchored at one child, the deltas from that child to its parent and to its sibling become more aligned, yielding higher cosine similarity and lower angles (\Cref{fig:hero_witness_overlap}b). This anchor-conditioned asymmetry suggests a directional rule: compare the overlap scores anchored at $A$ and $B$, and predict the lower-overlap anchor as the parent.

Based on this insight, we propose \textsc{Witness Overlap}, a training-free directional provenance test that uses a third same-family checkpoint $W$ as a witness (\Cref{fig:hero_witness_overlap}a--b). For a candidate pair $(A,B)$, we compute the overlap between the update directions from $A$ to $B$ and from $A$ to $W$, and compare it with the corresponding overlap when anchoring at $B$. We instantiate this idea with two prompt-free weight-space scores: Frobenius cosine over full weight-change directions and top-$k$ SVD subspace overlap over dominant update subspaces. We also provide a theoretical analysis that explains why this anchor-conditioned asymmetry can reveal directionality. Across open-weight LLM, VLM, and diffusion checkpoints, we evaluate \textsc{Witness Overlap} on three directional provenance tasks: identifying the root checkpoint within a same-family set, orienting a known parent--child pair, and distinguishing direct parent--child pairs from sibling pairs (\Cref{fig:hero_witness_overlap}c--e). We further define an additional task, single-chain ordering, in LLM families, where checkpoints are arranged along a sequential fine-tuning path using the same signal~(\Cref{fig:hero_witness_overlap}f). Across these four settings, \textsc{Witness Overlap} outperforms scalar directional baselines on the first two tasks and naturally extends to the latter two.

\paragraph{Contributions.}
\begin{enumerate}
\item We formulate \emph{family-conditional directional provenance} for open-weight model families. Given a same-family checkpoint set, we study four family-internal directional tasks: identifying the root checkpoint, orienting parent--child fine-tuning edges, distinguishing parent--child pairs from sibling pairs, and ordering checkpoints along a single fine-tuning chain.

\item We propose \textsc{Witness Overlap}, a relative weight-space test that uses a third same-family checkpoint as a witness. The method compares anchor-conditioned overlap scores and predicts the lower-overlap anchor as the parent. We instantiate it with two prompt-free, label-free, and training-free scores: Frobenius cosine over full weight-change directions and top-$k$ SVD subspace overlap over dominant low-rank update patterns. We also provide an analysis explaining why this anchor-conditioned asymmetry reveals directionality in model provenance setting.


\item We evaluate 176 LLM checkpoints from 16 families, 16 VLM checkpoints from four families, and 20 diffusion checkpoints from five families. With one same-family witness, \textsc{Witness Overlap} outperforms scalar baselines on root identification and parent--child orientation, exceeding 93\% LLM orientation accuracy and reaching 100\% LLM root accuracy. It also supports sibling discrimination with 0.898 AUROC and 82.4\% balanced accuracy on LLMs, and recovers all five tested LLM single-chain orders. Results are also strong beyond LLMs, including up to 100\% VLM root/orientation accuracy and 90\% diffusion parent--child orientation accuracy. Stress tests show robustness to weight noise and pruning, with the SVD variant more robust than Frobenius cosine.
\end{enumerate}

\section{Related Work}
\label{sec:related}

We organize related work around two provenance stages. Stage 1, family detection, asks whether checkpoints are related and constructs a same-family candidate set. Stage 2, family-internal lineage recovery, assumes such a same-family set is known and asks directional relationship between checkpoints.
\paragraph{Stage 1: family detection.}
One line of work uses
external provenance evidence from model cards \citep{mitchell2019modelcards}, repository
metadata and reuse records \citep{jiang2023ptmreuse,stalnaker2026huggingface}, and
supply-chain graphs \citep{rahman2025hugginggraph} to construct
same-family candidate sets and hypothesize lineage relations. However, these signals depend on release practices and may be incomplete, inconsistent, or missing. Another line of work uses model-derived evidence to detect related checkpoints. In white-box settings, this evidence
can come from representation fingerprints \citep{zhang2025reef,wu2026llmdna},
weight, gradient, or statistical-independence signals
\citep{zhu2025independence,wu2025tensorguard}, and activation or attention-pattern
fingerprints \citep{liu2026fnf,zhang2026attndiff}. In black-box settings, relatedness
can be inferred from behavioral evidence such as model-output hypothesis tests
\citep{nikolic2025mptllm}, prompt-injection fingerprints \citep{hu2025llmprint}, and
likelihood or generated-text traces that preserve training-order correlations
\citep{kuditipudi2025palimpsestic}. Recent results suggest
that this Stage-1 problem is already addressed well.
As a strong Stage-1 baseline, \citet{nikolic2025mptllm} uses black-box outputs and multiple hypothesis testing to detect model relatedness.
On the LLM models used in our study, it achieves 99.32\% precision and 91.25\% recall(\Cref{tab:mpt_stage1_zoo_subset}) when screening 160 true root-derived pairs against 2,400 negative child-root controls. However, relatedness
evidence alone does not determine the directional relationship within a family.
\paragraph{Stage 2: family lineage structure recovery.}
The closest prior work studies family-internal lineage structure recovery once models are known to be related. Neural Lineage predicts which candidate parent a fixed child model was fine-tuned from, with experiments mainly on visual models \citep{yu2024neurallineage}. Neural Phylogeny extends this line of work to detecting parent-child pairs and orienting their direction in a mixed set of models, using both learning-free norm-based cues and learned detectors, with experiments ranging from simple networks to Diffusion and LLaMA checkpoints \citep{yu2025neuralphylogeny}. MoTHer formulates unsupervised model-tree heritage recovery as a directed tree-recovery problem over weight-derived statistics, using weight distance for edge placement and kurtosis-based outlier monotonicity for edge direction \citep{horwitz2025mother}. 

In contrast, \textsc{Witness Overlap} targets a training-free Stage-2 setting inside an
already-identified model family. Rather than assigning each checkpoint a global scalar score, it uses
a third same-family checkpoint as a witness and derives direction from anchor-conditioned overlap
between weight deltas.
\section{Method: Witness Overlap}
\label{sec:method}

\subsection{Setting and notation}

Let $\F$ be a set of same-family open-weight checkpoints, consisting of a base model and post-trained descendants such as SFT or RLHF checkpoints. We assume that $\F$ is obtained from an upstream Stage-1 family-screening step, which we treat as a known outcomes from the first step rather than the target of this work.

Given such a same-family set, for each $M\in\F$ and corresponding matrix-valued tensor index $\ell$, let $\Wmat^{(\ell)}_M\in\R^{m\times n}$ denote the corresponding 2D weight tensor. We aggregate witness scores over an block set $\mathcal{B}$ of dense projection roles. In the LLM experiments, $\mathcal{B}$ denotes all the blocks except embeddings, lm head and normalization weights since our score targets shared layerwise update directions rather than architecture-specific input/output heads.
For VLM and Diffusion checks, $\mathcal{B}$ denotes the analogous architecture-specific projection blocks used by that family.

We study three Stage-2 tasks within a same-family set and one controlled sequential-chain task. \textbf{Root identification} predicts the base checkpoint $\hat r=\argmin_M\rootscore(M)$. \textbf{Parent-child orientation} assumes that $(A,B)$ is a direct parent-child pair and predicts which endpoint is the parent. \textbf{Pair-type discrimination} decides whether a same-family pair is a direct parent-child edge or a sibling pair that shares the same base. \textbf{Single-chain ordering} assumes a known base and orders successive SFT checkpoints in a controlled fine-tuning chain.

\subsection{Per-tensor anchor-conditioned statistics}

For an anchor $A$, witness $W$, and target $B$, define $\Delt^{(\ell)}_{AB}=\Wmat^{(\ell)}_B-\Wmat^{(\ell)}_A$ and $\Delt^{(\ell)}_{AW}=\Wmat^{(\ell)}_W-\Wmat^{(\ell)}_A$. We instantiate the witness statistic with two per-tensor scores.

\paragraph{Frobenius cosine.}
The first statistic treats each tensor update as a vector in Frobenius geometry:
\begin{equation}
\overlapfn_{\cos}^{(\ell)}(A;\,B,W)
=
\frac{\left\langle \Delt^{(\ell)}_{AB},\Delt^{(\ell)}_{AW}\right\rangle_F}
{\left\|\Delt^{(\ell)}_{AB}\right\|_F\left\|\Delt^{(\ell)}_{AW}\right\|_F}
\;\in\;[-1,1].
\label{eq:frob_cos}
\end{equation}
It measures whether the two outgoing deltas from the same anchor are co-oriented or anti-oriented in the full tensor space.

\paragraph{SVD top-$k$ subspace overlap.}
The second statistic compares dominant update subspaces. Compute a rank-$k$ truncated SVD, with default $k=16$, and let $\Vbasis^{(\ell)}_{AB},\Vbasis^{(\ell)}_{AW}\in\R^{n\times k}$ be the top-$k$ right-singular bases. The SVD overlap is
\begin{equation}
\overlapfn_{\mathrm{svd}}^{(\ell)}(A;\,B,W)
=
\frac{1}{k}\,\Frob{\Vbasis^{(\ell)\top}_{AW}\,\Vbasis^{(\ell)}_{AB}}^{2}
\;\in\;[0,1].
\label{eq:svd_overlap}
\end{equation}
This is the average squared cosine of the principal angles between the two top-$k$ right-singular subspaces. It extracts the principle components and smooths tensor-level noise by aggregating several dominant directions, making it more robust under weight pruning.

Both scores are anchor-asymmetric by construction: in general, $\overlapfn_{\star}^{(\ell)}(A;B,W)\neq\overlapfn_{\star}^{(\ell)}(B;A,W)$ for $\star\in\{\cos,\mathrm{svd}\}$. Unless otherwise specified, $\overlapfn$ denotes either of these per-tensor witness statistics, chosen according to the task.

\subsection{Weight aggregation and task decision rules}
\label{sec:method:aggregation}
For parent-child orientation and pair-type discrimination, the decision rules are defined with a single witness checkpoint $W\in\F\setminus\{A,B\}$. For root identification, the two non-anchor checkpoints play symmetric roles, so we average over all admissible witness pairs with the candidate root as the anchor. The effect of pooling additional witnesses for the parent-child tasks is analyzed separately in \Cref{sec:experiments:k_witness}.

Per-tensor scores are aggregated in two steps. First, we average over tensors within the same parameter block in $\mathcal{B}$. Second, we average over different parameter blocks in $\mathcal{B}$ to obtain a single triplet score
$\overlapfn(A;B,W)$. This score measures how similar the two outgoing deltas from anchor $A$ are, when the target endpoint is $B$ and the witness is $W$.

\paragraph{Root identification.}
For root identification, each candidate root $M$ is scored by averaging over unordered witness pairs with $M$ as the anchor:
\begin{align}
\rootscore(M)
&=
\frac{1}{\binom{|\F|-1}{2}}
\sum_{\{W_1,W_2\}\subset\F\setminus\{M\}}
\overlapfn(M;W_1,W_2),
\label{eq:root_score}\\
\hat r
&=\argmin_{M\in\F}\rootscore(M).
\label{eq:root_decision}
\end{align}
The root is the checkpoint whose outgoing deltas to other same-family checkpoints have the lowest average anchor-conditioned witness-pair overlap.

\paragraph{Parent-child orientation.}
Given a pair $(A,B)$ known to be a direct parent-child pair, we use the two endpoint anchor scores $s_A(A,B;W)=\overlapfn(A;B,W)$ and $s_B(A,B;W)=\overlapfn(B;A,W)$. The parent is predicted as the lower-overlap anchor:
\begin{equation}
\widehat{\mathrm{parent}}(A,B;W)
=
\argmin_{X\in\{A,B\}}s_X(A,B;W).
\label{eq:pc_direction_single_witness}
\end{equation}
The intuition is that a base or parent anchor has outgoing deltas pointing toward distinct descendants, whereas a child anchor produces deltas that share a reverse-update component.

\paragraph{Pair-type discrimination.}
For a general same-family model pair $(A,B)$, we distinguish a direct parent-child edge from a sibling pair using the endpoint score profile
$\bigl(s_A(A,B;W),\,s_B(A,B;W)\bigr)$. A parent-child pair has a low-high pattern, since the parent endpoint has low witness overlap and the child endpoint has higher witness overlap. A sibling pair has a high-high pattern, since both endpoints behave like child anchors relative to the witness. We therefore use the smaller endpoint score as the primary pair-type statistic, with the decision rule
\begin{equation}
\widehat{\mathrm{type}}(A,B;W)
=
\begin{cases}
\mathrm{parent\text{-}child}, & \min\{s_A(A,B;W),s_B(A,B;W)\}\le \tau_{\mathrm{type}},\\
\mathrm{sibling}, & \min\{s_A(A,B;W),s_B(A,B;W)\}> \tau_{\mathrm{type}}.
\end{cases}
\label{eq:pair_type_rule}
\end{equation}
The threshold $\tau_{\mathrm{type}}$ is selected by a validation set, or by holding out some model families for calibration, and then applied to unseen families.

\paragraph{Single-chain ordering.}
For a sequential fine-tuning chain with known base checkpoint $r$ and SFT checkpoints $\mathcal{C}=\{C_1,\ldots,C_T\}$, we anchor at each SFT checkpoint $A\in\mathcal{C}$, use the base $r$ as the second input of overlap, and try every other SFT checkpoint as the witness. A binary diagnostic counts high-overlap witnesses, $h(A)=\sum_{W\in\mathcal{C}\setminus\{A\}}\mathbf{1}[\overlapfn_{\cos}(A;\,r,W)>\tau_{\mathrm{chain}}]$. Practically, we use a simpler threshold-free chain score obtained by averaging these witness overlaps:
\begin{equation}
s_{\mathrm{chain}}(A)
=
\frac{1}{|\mathcal{C}|-1}
\sum_{W\in\mathcal{C}\setminus\{A\}}
\overlapfn_{\cos}(A;\,r,W).
\label{eq:single_chain_mean_score}
\end{equation}
We then order SFT checkpoints by increasing $s_{\mathrm{chain}}(A)$:
\begin{equation}
A \prec_{\mathrm{chain}} B
\quad\Longleftrightarrow\quad
s_{\mathrm{chain}}(A)<s_{\mathrm{chain}}(B).
\label{eq:single_chain_order}
\end{equation}
The intuition is that later checkpoints share reverse-update components with more earlier SFT witnesses, so later models in the chain have more high-overlap entries.

\subsection{Algorithm and properties}
\label{sec:method:properties}
The full procedure (Algorithm~\ref{alg:witness_overlap}, \Cref{app:algorithm}) iterates over ordered triplets $(A,B,W)\in\F^3$, computes the selected per-tensor statistic from \Cref{eq:frob_cos,eq:svd_overlap}, accumulates triplet scores, and aggregates as in \Cref{eq:pc_direction_single_witness,eq:pair_type_rule,eq:root_score}. An intuitive update-vector explanation for the single-chain ordering pattern is provided in \Cref{app:single_chain}.
\section{Proposition: When Parent Anchors Have Lower Witness Scores}
\label{sec:theory}

This section explains why witness scores are expected to be lower when the anchor is a parent than when the anchor is a child. We state the Frobenius-cosine result here because it gives a simple sign condition for the overlap between flattened weight updates to interpret the experiments. The SVD subspace extension and all derivations are deferred to \Cref{app:theory}.

\paragraph{Frobenius-cosine witness asymmetry.}
Let $\theta_m$ denote the collection of trainable weight parameters of checkpoint $m$. Let $p$ be the parent checkpoint and let $i,j$ be two descendants with $\theta_i=\theta_p+u_i$ and $\theta_j=\theta_p+u_j$, where $u_i$ and $u_j$ are their fine-tuning updates. For an anchor $a$ and two comparison checkpoints $b,c$, define the Frobenius witness cosine $C(a;b,c)=\langle \theta_b-\theta_a,\theta_c-\theta_a\rangle/(\|\theta_b-\theta_a\|\,\|\theta_c-\theta_a\|)$. The inner product and norm are computed after flattening the corresponding weight tensors. Let $\rho=\langle u_i,u_j\rangle/(\|u_i\|\,\|u_j\|)$ and $\lambda=\|u_i\|/\|u_j\|$. Thus, $\rho$ is the alignment between the two descendant updates, and $\lambda$ is their relative update norm.

\begin{proposition}[Frobenius-cosine witness asymmetry]
\label{prop:rank_one}
For nonzero, non-collinear updates $u_i,u_j$, the parent- and child-anchor Frobenius scores satisfy
\begin{equation}
C(p;i,j)=\rho,
\qquad
C(i;p,j)=\frac{\lambda-\rho}{\sqrt{\lambda^2+1-2\lambda\rho}}.
\label{eq:frobenius_cos_scores}
\end{equation}
Consequently,
\begin{equation}
C(i;p,j)>C(p;i,j)
\quad\Longleftrightarrow\quad
\rho<\frac{\lambda}{2}=\frac{\|u_i\|}{2\|u_j\|}.
\label{eq:frobenius_cos_inequality}
\end{equation}
\end{proposition}

Different children generally induce different update directions. Our experiments show that the observed alignments $\rho$ are typically small, which places real checkpoints in the regime where our theory \Cref{prop:rank_one} expects child anchors to have higher overlap scores.

\section{Experiments}
\label{sec:experiments}

We evaluate \textsc{Witness Overlap} on open-weight model families and compare it with the closest published directional baselines (\Cref{sec:experiments:main}) among the four tasks proposed in \Cref{sec:method:aggregation}. We report the two statistics from \Cref{sec:method}: the Frobenius cosine $\overlapfn_{\cos}$ and the SVD top-$k{=}16$ subspace overlap $\overlapfn_{\mathrm{svd}}$.

\subsection{Setup}
\label{sec:experiments:setup}

The main LLM evaluation contains $176$ checkpoints from $80\,{+}$ independent HuggingFace organizations~\citep{huggingfacehub}, grouped into $16$ same-family sets. Each family has one base checkpoint, and together they contain $130$ SFT descendants and $30$ RLHF descendants. To check whether the same anchor-role signal transfers beyond text-only LLMs, we also evaluate four VLM families and five Diffusion families. The full registries are listed in \Cref{app:rlhf_per_family}. The full LLM evaluation takes approximately $8$ GPU\,$\cdot$\,h on one $48$\,GB L40S. The VLM and Diffusion evaluations each take approximately $1$ GPU\,$\cdot$\,h on the same hardware. Our code will be hosted at https://anonymous.4open.science/r/witness-llm-DCA3/

\textbf{Lineage labels.} We build families from HuggingFace model-card meta
data and repository descriptions. For each base model, we collect compatible checkpoints from its HuggingFace fine-tune listings, ordered by download count, and include RLHF/alignment descendants when the description explicitly mentions DPO, GRPO, PPO, ORPO, or related alignment training. These labels reflect publicly declared HuggingFace lineage metadata, so our direct-edge claims should be read under that metadata-derived ground-truth convention.

\textbf{Baselines.} We run the closest published directional baselines on the same checkpoints. For Neural Phylogeny \citep{yu2025neuralphylogeny}, we use its parameter-norm endpoint rule: lower norm predicts parent, and lowest norm predicts root. We exclude its learning-based detector because it changes the setting from training-free Stage-2 comparison to supervised graph prediction with labeled train/test lineages. For MoTHer \citep{horwitz2025mother}, we evaluate its kurtosis direction component on fixed parent-child pairs and use its directed tree recovery for root identification.

\subsection{Main result}
\label{sec:experiments:main}

\begin{table}[t]
\centering
\small
\caption{\textbf{Root identification.} Accuracy of selecting the base checkpoint within each family. Relatedness methods such as REEF\citep{zhang2025reef}, LLM DNA\citep{wu2026llmdna}, and black-box provenance tests
\citep{nikolic2025mptllm}
are omitted because they use symmetric or non-orienting pairwise scores and therefore cannot directly identify the root within a same-family set.\textsc{Witness Overlap} chooses the checkpoint with the lowest mean overlap when used as an anchor. The L2 medoid selects the checkpoint with the lowest mean pairwise Euclidean distance. Neural Phylogeny uses its parameter-norm endpoint rule. MoTHer uses its kurtosis-based directed tree recovery.}
\label{tab:root_id}
\setlength{\tabcolsep}{8pt}
\begin{tabular}{p{5.4cm}ccc}
\toprule
Method & LLM & VLM & Diffusion \\
\midrule
\textbf{Witness Overlap, Frobenius cosine (ours)} & 93.8\% & \textbf{100.0\%} & \textbf{80.0\%} \\
\textbf{Witness Overlap, SVD top-$k{=}16$ (ours)} & \textbf{100.0\%} & \textbf{100.0\%} & \textbf{80.0\%} \\
L2 medoid & 81.3\% & 75.0\% & 60.0\% \\
Neural Phylogeny (norm) \citep{yu2025neuralphylogeny} & 18.8\% & 0.0\% & 20.0\% \\
MoTHer \citep{horwitz2025mother} & 18.8\% & 25.0\% & 60.0\% \\
\bottomrule
\end{tabular}
\end{table}

\begin{table}[t]
\centering
\small
\caption{\textbf{Parent-child direction accuracy at $\Nw{=}1$.} The LLM columns use the $16$-family open-weight LLM collection. VLM and Diffusion report four VLM families and five Diffusion families. Witness rows use one same-family witness per decision. Scalar baselines are witness-free.}
\label{tab:main_results}
\setlength{\tabcolsep}{3pt}
\resizebox{\textwidth}{!}{%
\begin{tabular}{lccccc}
\toprule
Method & LLM SFT & LLM RLHF & LLM Comb. & VLM SFT & Diffusion SFT \\
\midrule
\textbf{Witness Overlap, Frobenius cosine (ours)} & \textbf{96.2\%} & \textbf{91.4\%} & \textbf{95.3\%} & \textbf{100.0\%} & 85.0\% \\
\textbf{Witness Overlap, SVD top-$k{=}16$ (ours)} & 95.2\% & 86.4\% & 93.5\% & 87.5\% & \textbf{90.0\%} \\
Neural Phylogeny (norm) \citep{yu2025neuralphylogeny} & 79.2\% & 66.7\% & 76.9\% & 50.0\% & 50.0\% \\
MoTHer kurtosis \citep{horwitz2025mother} & 79.2\% & 53.3\% & 74.4\% & 68.8\% & 70.0\% \\
\bottomrule
\end{tabular}
}
\end{table}

\paragraph{Task1: Root identification}
\Cref{tab:root_id} reports root identification accuracy across LLM, VLM, and Diffusion families. The lower-overlap rule identifies the root by selecting the checkpoint with the lowest mean witness overlap. The SVD-based metric identifies $16/16$ LLM, $4/4$ VLM roots, and $4/5$ Diffusion roots. Frobenius cosine identifies all VLM roots and $4/5$ Diffusion roots, but misses one LLM root. We also evaluate an L2 medoid that selects the checkpoint with the lowest mean Euclidean distance to the other family members. It identifies $13/16$ LLM, $3/4$ VLM, and $3/5$ Diffusion roots, below the best Witness Overlap result in every domain. L2 uses distance magnitude only, so a descendant with a small update can appear more central than the true root. Witness Overlap instead uses directional relationships between updates. The remaining endpoint-only baselines are substantially weaker. On LLMs, both Neural Phylogeny's norm rule and MoTHer identify only $18.8\%$ of roots. Across VLM and Diffusion families, Neural Phylogeny reaches $0.0\%$ and $20.0\%$, while MoTHer reaches $25.0\%$ and $60.0\%$, respectively.

\begin{figure}[t]
    \centering
    \begin{subfigure}[t]{0.32\textwidth}\centering
        \includegraphics[width=\linewidth]{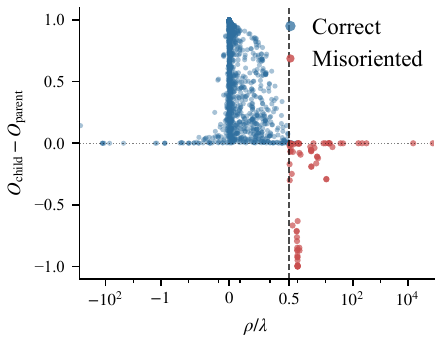}
        \caption{Frobenius-cosine margin vs $\rho/\lambda$.}
        \label{fig:cos_margin_vs_rho}
    \end{subfigure}\hfill
    \begin{subfigure}[t]{0.32\textwidth}\centering
        \includegraphics[width=\linewidth]{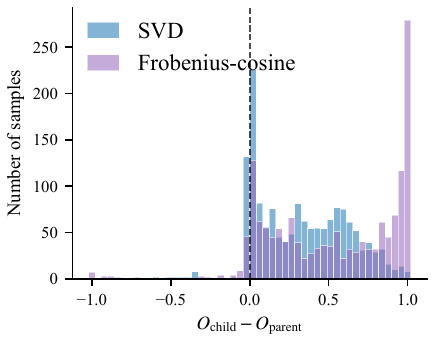}
        \caption{Margin distribution.}
        \label{fig:margin_distribution}
    \end{subfigure}\hfill
    \begin{subfigure}[t]{0.32\textwidth}\centering
        \includegraphics[width=\linewidth]{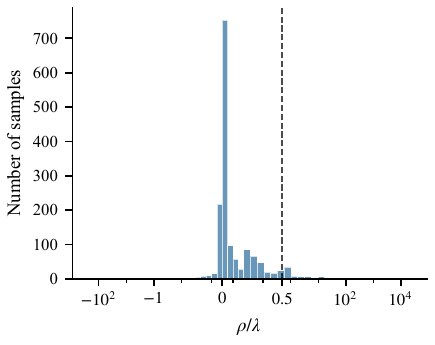}
        \caption{$\rho/\lambda$ distribution.}
        \label{fig:rho_lambda_distribution}
    \end{subfigure}
    \caption{\textbf{Single-witness Frobenius-cosine margin and $\rho/\lambda$ scatter plot and distribution.} Panel (a) plots Frobenius-cosine margin at $\Nw{=}1$ against $\rho/\lambda$ on a symlog scale, and the dashed line marks the sufficient-condition boundary $\rho/\lambda=1/2$ from \Cref{prop:rank_one}. Panel (b) shows the single-witness margin distribution over all LLM triplets. SVD has $1442/1542$ positive margins with median $+0.31$, and Frobenius cosine has $1469/1542$ positive margins with median $+0.57$. Panel (c) shows the empirical distribution of $\rho/\lambda$ for the same Frobenius-cosine triplets, with most values below $0.5$.}
    \label{fig:margins_and_rank_one}
\end{figure}

\paragraph{Task2: Parent-child direction}
\Cref{tab:main_results} reports single-witness ($\Nw{=}1$) parent-child direction accuracy, where one additional same-family checkpoint is used as the witness for each decision. On the LLM collection, $\overlapfn_{\cos}$ reaches $95.3\%$ direction accuracy and $\overlapfn_{\mathrm{svd}}$ reaches $93.5\%$. Evaluations on the VLM collection show the same qualitative pattern: Frobenius cosine correctly orients all VLM decisions, while the SVD-based metric reaches 87.5\% accuracy. On Diffusion families, SVD reaches $90.0\%$ accuracy, and Frobenius cosine reaches $85.0\%$. Endpoint-only baselines are weaker. On LLMs, Neural Phylogeny's norm rule reaches $76.9\%$ over all parent-child accuracy, while MoTHer reaches $74.4\%$ accuracy. We further examine why the direction rule succeeds by looking at single-witness margin, which is child overlap minus parent overlap. \Cref{fig:cos_margin_vs_rho} shows that most LLM triplets have positive Frobenius-cosine margins and lie left of the $\rho/\lambda=1/2$ boundary. All $73$ Frobenius-cosine failures lie to its right, broadly consistent with the trend predicted by \Cref{prop:rank_one}. \Cref{fig:margin_distribution} compares the full margin distributions for Frobenius cosine and SVD top-$k$ overlap. Both distributions are concentrated on positive margins, empirically supporting that \Cref{eq:pc_direction_single_witness} correctly orients most LLM triplets for both overlap statistics.

\paragraph{Task3: Pair-type discrimination(sibling vs.\ parent-child)}
\begin{table}[t]
\centering
\small
\caption{\textbf{Pairwise-type discrimination at $\Nw{=}1$, leave-one-family-out.} A logistic regressor on $\min\{s_A(A,B;W),s_B(A,B;W)\}$ distinguishes parent--child from sibling pairs using one witness. Thresholds are selected only on training families. Bal. acc. denotes balanced accuracy.}
\label{tab:pairwise_discrim}
\setlength{\tabcolsep}{4pt}
\begin{tabular}{l cc cc cc}
\toprule
Method & \multicolumn{2}{c}{LLM} & \multicolumn{2}{c}{VLM} & \multicolumn{2}{c}{Diffusion} \\
\cmidrule(lr){2-3}\cmidrule(lr){4-5}\cmidrule(lr){6-7}
& AUROC & Bal.\ acc & AUROC & Bal.\ acc & AUROC & Bal.\ acc \\
\midrule
Frobenius cosine & \textbf{0.898} & \textbf{82.4\%} & \textbf{0.903} & \textbf{79.9\%} & \textbf{0.855} & \textbf{80.3\%} \\
SVD top-$k{=}16$ & 0.780 & 68.6\% & 0.715 & 54.9\% & \textbf{0.855} & 53.3\% \\
\bottomrule
\end{tabular}
\end{table}

Beyond orienting known edges, we ask whether one witness can distinguish parent-child pairs from sibling pairs. Following \Cref{eq:pair_type_rule}, \Cref{tab:pairwise_discrim} uses the smaller endpoint-anchor score $\min\{s_A(A,B;W),s_B(A,B;W)\}$ as the scalar input to a leave-one-family-out logistic regressor, with the binary label indicating whether (A,B) is a parent–child or sibling pair. The threshold is calibrated on training families and evaluated on the held-out family. Pair-type discrimination is therefore not training-free. On LLMs, Frobenius cosine reaches $0.898$ AUROC and $82.4\%$ balanced accuracy, while SVD reaches $0.780$ AUROC and $68.6\%$ balanced accuracy. Frobenius cosine also performs best on the VLM and Diffusion checks, reaching $79.9\%$ and $80.3\%$ balanced accuracy. These results are lower than the direction and root-identification accuracies, as pair typing requires an additional cross-family decision boundary.

\begin{figure}[t]
    \centering
    \scalebox{0.9}{
    \begin{minipage}{\textwidth}
    \centering
    \begin{subfigure}[t]{0.48\textwidth}\centering
        \includegraphics[width=\linewidth]{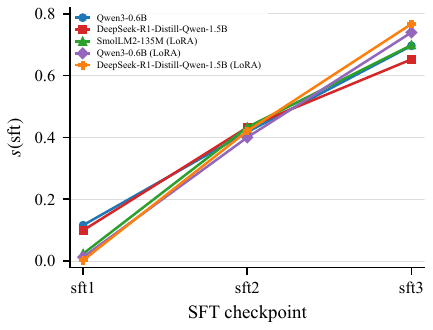}
        \caption{Single-chain ordering.}
        \label{fig:single_chain_s_scores}
    \end{subfigure}\hfill
    \begin{subfigure}[t]{0.48\textwidth}\centering
        \includegraphics[width=\linewidth]{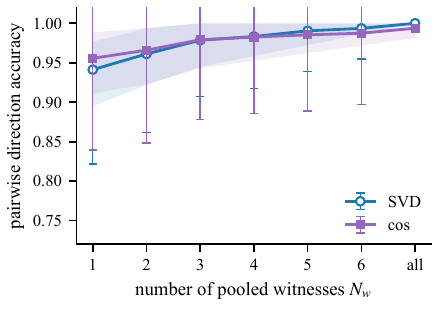}
        \caption{LLM witness averaging.}
        \label{fig:kwitness_sweep_llm}
    \end{subfigure}
    \caption{\textbf{Single-chain ordering and witness averaging.} Panel (a) plots the mean witness-overlap score $s(\mathrm{sft})$ for each checkpoint in the controlled chain, where larger scores indicate later checkpoints. Panel (b) tests whether pooling same-family witnesses stabilizes LLM parent-child direction accuracy.}
    \label{fig:chain_and_kwitness}
    \end{minipage}
    }
\end{figure}

\vspace{-2mm}
\paragraph{Task 4: Single-chain ordering.}
\label{sec:experiments:single_chain}
Beyond released same-family collections, we test controlled lineages that we fine-tune ourselves. For each base model, we run continued supervised fine-tuning, using either full-weight fine-tuning or LoRA, for one epoch on each dataset in a fixed order: MMLU ~\citep{hendrycks2021measuring} produces $sft1$, UltraChat ~\citep{ding-etal-2023-enhancing} continues from $sft1$ to produce $sft2$, and PubMedQA~\citep{jin-etal-2019-pubmedqa} continues from $sft2$ to produce $sft3$. Following \Cref{eq:single_chain_mean_score}, for each SFT checkpoint $A$ we use $A$ as the anchor, compare its direction to the base with its directions to the other SFT checkpoints, and define $s(A)$ as the average Frobenius-cosine witness overlap.
As shown in \Cref{fig:single_chain_s_scores}, the mean score increases from $sft1$ to $sft2$ to $sft3$ across all five controlled chains. 

\vspace{-2mm}
\paragraph{Anchor-role asymmetry.}

\Cref{tab:anchor_overlap} confirms the expected anchor-role asymmetry across LLM, VLM, and Diffusion families. Averaged over all witnesses for each LLM parent-child pair, the mean parent-anchor and child-anchor scores are $0.072$ and $0.396$ for $\overlapfn_{\mathrm{svd}}$, and $0.023$ and $0.553$ for $\overlapfn_{\cos}$. We can see the child-anchor overlap scores is far larger than parent-anchor overlap scores. The same ordering also holds on VLMs and Diffusion families, with architecture-dependent scale. Across LLM triplets, the median $\rho/\lambda$ is $8.08\times 10^{-4}$ from \Cref{app:family_geometry}, far below the $1/2$ boundary in \Cref{prop:rank_one}, consistent with the setting where child and witness updates have weak direct alignment and the asymmetry should be visible.

\vspace{-2mm}
\subsection{\texorpdfstring{$\Nw$}{Nw}-witness ablation}
\label{sec:experiments:k_witness}
\label{sec:experiments:margins}
\vspace{1mm}
\vspace{-3mm}

To evaluate how witness pooling affects the stability of the directional signal, for the LLM evaluation, we average the scores from $\Nw$ randomly selected same-family witnesses for each observed parent-child pair and then decide the parent–child direction using these averaged scores. Thus $\Nw{=}1$ is the single-witness setting, while $\Nw{=}\text{all}$ uses every eligible witness in that family. This ablation is a stability check rather than the main directionality claim. SVD top-$k{=}16$ rises from $94.1\%$ at $\Nw{=}1$ to $97.9\%$ at $\Nw{=}3$ and $100.0\%$ at $\Nw{=}\text{all}$. Frobenius cosine moves from $95.6\%$ to $97.9\%$ and $99.4\%$, respectively. The shaded bands in \Cref{fig:kwitness_sweep_llm} are family-level bootstrap $95\%$ confidence intervals, obtained by resampling families and recomputing the accuracy. On VLM and Diffusion families, averaging over all eligible witnesses shows the same trend: pooling witnesses stabilizes the directional signal beyond the single-witness setting. These results are shown in \Cref{fig:kwitness_sweep_crossdomain}.

\vspace{-2mm}
\subsection{Failure case analysis}
\label{sec:experiments:failure}
\vspace{-2mm}
To understand when the single-witness directional signal breaks down, we examine the one-witness LLM triplets that are misoriented: SVD top-$k{=}16$ misses $100/1542$, and Frobenius cosine misses $73/1542$. These failures are tail cases of the geometry in \Cref{app:family_geometry}: high child-witness alignment $\rho=C(p;i,j)$ and large witness-to-child scale $\eta=\|u_j\|/\|u_i\|=1/\lambda$ both push triplets toward violating $\rho<\lambda/2$ in \eqref{eq:frobenius_cos_inequality}. The failures are most frequent in SmolLM2-135M, where Frobenius cosine fails on $34/182$ triplets. \Cref{tab:cos_failure_cases} lists all Frobenius-cosine failures, and \Cref{fig:kwitness_sweep_llm} shows that averaging witnesses reduces the influence of these tail cases.

\vspace{-2mm}
\subsection{Noise perturbation and pruning robustness tests}
\label{sec:experiments:robustness}
\vspace{-2mm}
To test whether the directional signal is robust to weight-space perturbations, we evaluate it under noise perturbation and pruning stress tests. \Cref{app:child_noise_stress} adds Gaussian noise directly to the child checkpoint on GPT-2 and TinyLlama, with the noise magnitude set relative to the Frobenius norm of the clean parent-to-child update. Direction accuracy does not decrease across the tested noise levels. \Cref{app:magnitude_pruning_stress} further applies shape-preserving magnitude pruning to GPT-2 projection tensors in child-only, half-descendant, and all-descendant settings. Child-only pruning remains stable through 50\% sparsity, while pruning descendants is a stronger perturbation whose side effects are reduced by averaging more witnesses in milder settings. In the all-descendant setting, SVD is more robust than Frobenius cosine at moderate sparsity. Together, these tests suggest that the anchor-role asymmetry is robust to child-side noise and moderately sparse pruning, but degrades under severe pruning applied to all descendants.

\subsection{Same-family model merges}
\label{sec:experiments:same_family_merges}
We test merges of two descendants that share the same family root. We use TIES~\citep{yadav2023ties} and SLERP~\citep{shoemake1985animating} with $0.50/0.50$ and $0.25/0.75$ merge weights. With one witness, root--merge direction accuracy is $96.88\%$--$97.03\%$ for TIES and $95.70\%$--$96.56\%$ for SLERP across cosine and SVD. With all witnesses, both metrics reach $100\%$ for TIES. SVD also reaches $100\%$ for SLERP, while cosine reaches $97.50\%$--$98.13\%$. The method therefore usually places a same-family merge on the descendant side of its shared root. This experiment does not recover the models that contributed to the merge. \Cref{app:same_family_merges} gives the full setup and results, including the root-identification control.

\subsection{Quantization robustness test}
\label{sec:experiments:quantization}
We evaluate symmetric per-tensor signed INT8 quantize--dequantize in two settings. When every checkpoint in three complete LLM families is quantized, \textsc{Witness Overlap} identifies all $3/3$ roots and orients all $37/37$ edges with all witnesses. Single-witness direction accuracy is $423/436$ ($97.0\%$) for cosine and $406/436$ ($93.1\%$) for SVD. We also quantize only the queried child in five LLM families while keeping the parent and witnesses clean. Both metrics then reach $794/800$ ($99.25\%$) with one witness and $65/65$ with all witnesses. These results apply when checkpoints are available in a common dense parameterization. \Cref{app:int8_quantization} gives the full results and explains the apparent improvement in the child-only setting.

\vspace{-3mm}
\section{Discussion and Limitations}
\vspace{-3mm}
\label{sec:discussion}
\paragraph{Summary of capabilities.}
\textsc{Witness Overlap} is a simple weight-space method that compares anchor-conditioned update overlap using a same-family witness. Root identification reaches up to $100\%$ on LLMs and VLMs and $80\%$ on Diffusion models. With one witness, parent--child direction reaches $95.3\%$ on LLMs, $100\%$ on VLMs, and $90\%$ on Diffusion models. It also recovers all five tested single-chain orders. Root identification, parent--child direction, and single-chain ordering are training-free. Pair-type discrimination instead requires a calibrated threshold. The method transfers across the evaluated LLM, VLM, and Diffusion architectures, including U-Net-based diffusion checkpoints, as long as each family has a shared aligned parameter subspace.

\paragraph{Why the signal appears.}
Modern models have millions to billions of weights and are fine-tuned on varied tasks and datasets. We observe that their updates are often weakly aligned. This resembles concentration in high-dimensional space and makes the sufficient condition in \Cref{prop:rank_one} easier to satisfy.

\paragraph{Required inputs.}
The method assumes that Stage 1 provides an approximately correct same-family set. It also needs at least one additional same-family checkpoint as a witness and should abstain when none exists. A stress test with one Llama-3-8B or Mistral-7B false positive retains $22/22$ root predictions and $220/220$ all-witness directions. Single-witness accuracy remains $97.50\%$ for cosine and $96.09\%$ for SVD. This shows tolerance to limited contamination, but not to arbitrary Stage-1 errors (\Cref{app:stage1_false_positives}).

\paragraph{Quantization and merging.}
The method requires a shared, aligned parameter space. Under the evaluated INT8 rule, it retains $3/3$ root predictions and $37/37$ all-witness directions when all checkpoints are quantized. It also retains $65/65$ all-witness directions when only the child is quantized (\Cref{app:int8_quantization}). For same-family TIES and SLERP merges, root-to-merge direction also remains stable (\Cref{app:same_family_merges}). Recovering all merge contributors is outside our task. Cross-family merges violate the same-family premise and need not follow our shared-root update geometry, so we make no directional claim for them.

\paragraph{Remaining limitations.}
The signal can weaken when a child update is much smaller than a witness update or when two descendants have highly aligned updates (\Cref{sec:experiments:failure}). Pair-type discrimination is also less robust than root identification and direction. Its best balanced accuracy is below $90\%$. Finally, most experiments use direct base-to-descendant releases. The single-chain results are encouraging, but complex lineage reconstruction and adversarial tampering remain outside the validated setting (\Cref{sec:experiments:single_chain}).




\FloatBarrier
\bibliographystyle{plainnat}
\bibliography{references}
\FloatBarrier
\appendix
\newpage
\section{Evidence for Stage-1 Screening}
\label{app:stage1_screening}
We apply black-box outputs and multiple hypothesis testing \citep{nikolic2025mptllm} to the 176-checkpoint LLM subset used in our evaluation. This subset contains 160 true root-derived pairs and 2,400 unrelated child-root control pairs. \Cref{tab:mpt_stage1_zoo_subset} reports the resulting Stage-1 derived-model screening performance.

\begin{table}[H]
\centering
\caption{
Black-box outputs and multiple hypothesis testing \citep{nikolic2025mptllm}'s performance on our 176 LLM models.
The evaluation contains 160 true root-derived pairs and 2,400 unrelated child-root controls
(TP=146, FN=14, FP=1, TN=2,399).
}
\label{tab:mpt_stage1_zoo_subset}
\small
\begin{tabular}{lc}
\toprule
Metric & Value \\
\midrule
Precision & 99.32\% \\
Recall & 91.25\% \\
Specificity & 99.96\% \\
F1 & 95.11\% \\
Accuracy & 99.41\% \\
\bottomrule
\end{tabular}
\end{table}

\section{Stage-1 False-Positive Stress Test}
\label{app:stage1_false_positives}

We test whether one architecture-compatible false positive from Stage 1 disrupts the Stage-2 decisions. The Llama-3-8B and Mistral-7B families each contain one root and ten descendants. We inject each of the $11$ Llama-3-8B checkpoints into the complete Mistral-7B family, then repeat the process in the other direction. This produces $22$ contaminated pools. In each pool, we evaluate the ten true root-to-descendant edges, giving $220$ edge--contaminant conditions.

\begin{table}[H]
\centering
\footnotesize
\caption{\textbf{Stage-1 false-positive stress test.} One architecture-compatible checkpoint from the Llama-3-8B or Mistral-7B family is injected into the other family. Root ID is evaluated over the $22$ contaminated pools. Direction is evaluated over $220$ true edges with ten eligible witness choices per edge.}
\label{tab:stage1_false_positives}
\setlength{\tabcolsep}{6pt}
\begin{tabular}{lccc}
\toprule
Metric & Root ID & Random-one direction & All-witness direction \\
\midrule
Cosine & $22/22\ (100\%)$ & $2{,}145/2{,}200\ (97.50\%)$ & $220/220\ (100\%)$ \\
SVD & $22/22\ (100\%)$ & $2{,}114/2{,}200\ (96.09\%)$ & $220/220\ (100\%)$ \\
\bottomrule
\end{tabular}
\end{table}

The random-one setting evaluates each of the ten eligible witnesses in turn, giving $2{,}200$ direction decisions per metric. The all-witness setting averages the ten scores for each true edge. Despite the injected unrelated checkpoint, both metrics identify the true root in all $22$ pools and orient all $220$ true edges after witness aggregation. This stress test covers one false positive between two architecture-compatible LLM families. It does not establish robustness to arbitrary contamination or to pools containing several unrelated lineages.

\FloatBarrier

\section{Derivations for Witness Asymmetry}
\label{app:theory}
\label{app:proofs}

\subsection{Frobenius-cosine witness asymmetry}
\label{app:frobenius_cosine_proof}

Let $\theta_i=\theta_p+u_i$ and $\theta_j=\theta_p+u_j$. For any anchor $a$ and endpoints $b,c$, define
\[
C(a;b,c)
=
\frac{\langle \theta_b-\theta_a,\theta_c-\theta_a\rangle}
{\|\theta_b-\theta_a\|\,\|\theta_c-\theta_a\|}.
\]
Write
\[
\lambda=\frac{\|u_i\|}{\|u_j\|},
\qquad
\rho=\frac{\langle u_i,u_j\rangle}{\|u_i\|\,\|u_j\|}.
\]

For the parent anchor $p$, the two outgoing deltas are
\[
\theta_i-\theta_p=u_i,
\qquad
\theta_j-\theta_p=u_j.
\]
Therefore
\[
C(p;i,j)
=
\frac{\langle u_i,u_j\rangle}{\|u_i\|\,\|u_j\|}
=
\rho.
\]

For the child anchor $i$, the two outgoing deltas are
\[
\theta_p-\theta_i=-u_i,
\qquad
\theta_j-\theta_i=u_j-u_i.
\]
Thus
\[
C(i;p,j)
=
\frac{\langle -u_i,u_j-u_i\rangle}{\|u_i\|\,\|u_j-u_i\|}.
\]
The numerator is
\[
\langle -u_i,u_j-u_i\rangle
=
\|u_i\|^2-\langle u_i,u_j\rangle
=
\|u_i\|\,\|u_j\|(\lambda-\rho),
\]
and the second norm in the denominator is
\[
\|u_j-u_i\|
=
\|u_j\|\sqrt{1+\lambda^2-2\lambda\rho}.
\]
Hence
\[
C(i;p,j)
=
\frac{\lambda-\rho}{\sqrt{\lambda^2+1-2\lambda\rho}}.
\]

For $\rho<0$, the child-anchor score is positive while $C(p;i,j)=\rho$ is negative. For $0\le \rho<1$, compare the two explicit expressions:
\[
C(i;p,j)>C(p;i,j)
\quad\Longleftrightarrow\quad
\frac{\lambda-\rho}{\sqrt{\lambda^2+1-2\lambda\rho}}>\rho.
\]
This inequality holds exactly when
\[
(\lambda-\rho)^2>\rho^2(\lambda^2+1-2\lambda\rho)
\quad\Longleftrightarrow\quad
\lambda(1-\rho^2)(\lambda-2\rho)>0.
\]
Because $\lambda>0$ and $1-\rho^2>0$, this is equivalent to $\rho<\lambda/2$. In the non-collinear case $\rho<1$,
\[
C(i;p,j)>C(p;i,j)
\quad\Longleftrightarrow\quad
\rho<\frac{\lambda}{2}
=
\frac{\|u_i\|}{2\|u_j\|}.
\]
The collinear case $\rho=1$ is degenerate for this comparison and is outside the low-alignment regime used in the experiments.

\subsection{SVD subspace extension}
\label{app:svd_subspace_proof}

The SVD statistic applies the same witness idea to dominant right singular subspaces. For one aligned weight matrix, let $X_i=\Wmat_i-\Wmat_p$ and $X_j=\Wmat_j-\Wmat_p$. We use the same effective rank $k$ as the SVD overlap in \Cref{eq:svd_overlap}. Under the shared-left-subspace SVD model,
\[
X_i=L\Sigma_i V_i^{\top}
=\sum_{t=1}^{k}\sigma_{i,t}\ell_t v_{i,t}^{\top},
\qquad
X_j=L\Sigma_j V_j^{\top}
=\sum_{t=1}^{k}\sigma_{j,t}\ell_t v_{j,t}^{\top},
\]
where the left singular vectors $\ell_t$ are orthonormal and shared, while the singular values $\sigma_{i,t},\sigma_{j,t}>0$ may differ. Let $V_i=[v_{i,1},\ldots,v_{i,k}]$ and $V_j=[v_{j,1},\ldots,v_{j,k}]$. We choose the right bases as principal-vector bases for the two right singular subspaces, meaning
\[
V_i^{\top}V_j=\diag(\rho_1,\ldots,\rho_k),
\qquad
0\le \rho_t\le 1.
\]
Equivalently, $\rho_t=v_{i,t}^{\top}v_{j,t}$ and $v_{i,t}^{\top}v_{j,u}=0$ for $t\ne u$. Thus $\rho_t$ is the $t$-th principal cosine between the right subspaces of $X_i$ and $X_j$.

\begin{proposition}[SVD-subspace witness asymmetry]
\label{prop:svd_subspace}
Under the model above, if each paired right-subspace direction satisfies
$\rho_t<\min\{1,\sigma_{i,t}/(2\sigma_{j,t})\}$, then the child-anchor SVD overlap is larger than the parent-anchor SVD overlap: $O_i>O_p$.
\end{proposition}

For the parent anchor, the two right-singular bases are $V_i$ and $V_j$. Therefore the SVD overlap has exactly the same form as \Cref{eq:svd_overlap}:
\[
O_p
=
\frac{1}{k}\left\|V_i^{\top}V_j\right\|_F^2
=
\frac{1}{k}\sum_{t=1}^{k}\rho_t^2.
\]
For the child anchor, the parent-directed delta is
\[
\Delt_{i p}
=\Wmat_p-\Wmat_i
=-X_i
=-\sum_{t=1}^{k}\sigma_{i,t}\ell_t v_{i,t}^{\top}.
\]
Thus $\Delt_{i p}$ has right-singular basis $V_i$, because changing the sign of a matrix does not change its singular subspaces. The witness-directed delta is
\[
\begin{aligned}
\Delt_{i j}
&=\Wmat_j-\Wmat_i \\
&=(\Wmat_j-\Wmat_p)-(\Wmat_i-\Wmat_p) \\
&=X_j-X_i \\
&=\sum_{t=1}^{k}\sigma_{j,t}\ell_t v_{j,t}^{\top}
-\sum_{t=1}^{k}\sigma_{i,t}\ell_t v_{i,t}^{\top} \\
&=\sum_{t=1}^{k}\ell_t
\left(\sigma_{j,t}v_{j,t}-\sigma_{i,t}v_{i,t}\right)^{\top}
=\sum_{t=1}^{k}\ell_t z_t^{\top},
\end{aligned}
\]
where
\[
z_t=\sigma_{j,t}v_{j,t}-\sigma_{i,t}v_{i,t}.
\]
The vector $z_t$ is the unnormalized right factor paired with the shared left vector $\ell_t$ in the child-to-witness delta. The principal-vector choice makes these right factors mutually orthogonal. For $t\ne u$, $z_t^{\top}z_u=0$. Their norms are
\[
\|z_t\|^2=\sigma_{i,t}^2+\sigma_{j,t}^2-2\sigma_{i,t}\sigma_{j,t}\rho_t,
\]
so the normalized vectors $\hat z_t=z_t/\|z_t\|$ are the corresponding right singular directions of $\Delt_{i j}$, up to signs. Let
\[
Z=[\hat z_1,\ldots,\hat z_k].
\]
The child-anchor SVD overlap is therefore
\[
O_i
=
\frac{1}{k}\left\|V_i^{\top}Z\right\|_F^2.
\]
Since $V_i^\top Z$ is diagonal under the principal-vector pairing, its $t$-th diagonal contribution is
\[
c_t
:=
\left\langle v_{i,t},\hat z_t\right\rangle^2
=
\frac{(\sigma_{i,t}-\sigma_{j,t}\rho_t)^2}{\sigma_{i,t}^2+\sigma_{j,t}^2-2\sigma_{i,t}\sigma_{j,t}\rho_t}
\,.
\]
Therefore
\[
O_i=\frac{1}{k}\sum_{t=1}^{k}c_t,
\]
and hence
\[
O_i-O_p
=
\frac{1}{k}
\left(
\left\|V_i^{\top}Z\right\|_F^2
-
\left\|V_i^{\top}V_j\right\|_F^2
\right)
=
\frac{1}{k}\sum_{t=1}^{k}\left(c_t-\rho_t^2\right).
\]
It remains to compare each paired contribution. For $\rho_t<1$,
\[
c_t>\rho_t^2
\quad\Longleftrightarrow\quad
(\sigma_{i,t}-\sigma_{j,t}\rho_t)^2>\rho_t^2(\sigma_{i,t}^2+\sigma_{j,t}^2-2\sigma_{i,t}\sigma_{j,t}\rho_t)
\]
\[
\Longleftrightarrow\quad
\sigma_{i,t}(1-\rho_t^2)(\sigma_{i,t}-2\sigma_{j,t}\rho_t)>0.
\]
Since $\sigma_{i,t}>0$ and $1-\rho_t^2>0$ in the non-degenerate case $\rho_t<1$, this is equivalent to
\[
\rho_t<\frac{\sigma_{i,t}}{2\sigma_{j,t}}.
\]
Thus $c_t>\rho_t^2$ whenever $\rho_t<\min\{1,\sigma_{i,t}/(2\sigma_{j,t})\}$. If this holds for every paired direction, then $O_i>O_p$.

\section{Algorithm}
\label{app:algorithm}

\begin{algorithm}[H]
\caption{\textsc{Witness Overlap}}
\label{alg:witness_overlap}
\begin{algorithmic}[1]
\Require Same-family set $\F$, shared 2D tensor set $T$, statistic choice $\star\in\{\cos,\mathrm{svd}\}$, SVD rank $k$ (used only if $\star=\mathrm{svd}$), pair-type threshold $\tau_{\mathrm{type}}$
\Ensure Root prediction $\hat r$, parent-child orientations, pair-type predictions
\For{each anchor $A\in\F$, target $B\in\F\setminus\{A\}$, witness $W\in\F\setminus\{A,B\}$}
  \For{each tensor $\ell\in T$}
    \State $\Delt^{(\ell)}_{AB}\gets\Wmat^{(\ell)}_B-\Wmat^{(\ell)}_A$;\quad $\Delt^{(\ell)}_{AW}\gets\Wmat^{(\ell)}_W-\Wmat^{(\ell)}_A$
    \If{$\star=\cos$}
      \State $s_\ell\gets\langle\Delt^{(\ell)}_{AB},\Delt^{(\ell)}_{AW}\rangle_F /\bigl(\Frob{\Delt^{(\ell)}_{AB}}\Frob{\Delt^{(\ell)}_{AW}}\bigr)$ \Comment{\eqref{eq:frob_cos}}
    \Else
      \State $\Vbasis^{(\ell)}_{AB},\Vbasis^{(\ell)}_{AW}\gets$ top-$k$ right singular vectors of $\Delt^{(\ell)}_{AB},\Delt^{(\ell)}_{AW}$
      \State $s_\ell\gets\Frob{\Vbasis^{(\ell)\top}_{AW}\Vbasis^{(\ell)}_{AB}}^{2}/k$ \Comment{\eqref{eq:svd_overlap}}
    \EndIf
  \EndFor
  \State $\overlapfn(A;B,W)\gets$ block-aware mean of $\{s_\ell\}$: average within each block role then across block roles
\EndFor
\State \textbf{Root:} $\rootscore(M)\gets\frac{1}{\binom{|\F|-1}{2}}\!\sum_{\{W_1,W_2\}\subset\F\setminus\{M\}}\!\overlapfn(M;W_1,W_2)$;\quad $\hat r\gets\argmin_{M\in\F}\rootscore(M)$ \Comment{\eqref{eq:root_score}}
\For{each pair $(A,B)\subset\F$}
  \State $s_A\gets\mathrm{mean}_W\,\overlapfn(A;B,W)$; \quad $s_B\gets\mathrm{mean}_W\,\overlapfn(B;A,W)$
  \State \textbf{Parent-child orientation:} $\widehat{\mathrm{parent}}(A,B)\gets\argmin_{X\in\{A,B\}}s_X$ \Comment{\eqref{eq:pc_direction_single_witness}}
  \State \textbf{Pair-type:} $\widehat{\mathrm{type}}(A,B)\gets\mathrm{parent\text{-}child}$ if $\min\{s_A,s_B\}\le\tau_{\mathrm{type}}$, else $\mathrm{sibling}$
\EndFor
\end{algorithmic}
\end{algorithm}

\section{Single-Chain Ordering Details}
\label{app:single_chain}

For the single-chain ordering result in \Cref{sec:experiments:single_chain}, we use controlled chains that we fine-tune ourselves. We then compute the Frobenius cosine with the base checkpoint as target. For each base model, the three checkpoints are obtained by continued supervised fine-tuning for one epoch per dataset: $sft1$ is stage0\_mmlu, fine-tuned from the base on MMLU; $sft2$ is stage1\_ultrachat, continued from $sft1$ on UltraChat; and $sft3$ is stage2\_pubmedqa, continued from $sft2$ on PubMedQA. Write the resulting chain as
\[
\theta_1=\theta_0+u_1,\qquad
\theta_2=\theta_0+u_1+u_2,\qquad
\theta_3=\theta_0+u_1+u_2+u_3,
\]
where $\theta_0$ is the base and $\theta_t$ denotes $\mathrm{sft}_t$. For anchor $sft1$, the base direction is $-u_1$, while the two witness directions are $u_2$ and $u_2+u_3$, giving the low/low pattern. For anchor $sft2$, the base direction is $-(u_1+u_2)$; the $sft1$ witness direction is $-u_2$, which shares the reverse $u_2$ component, while the $sft3$ direction is $u_3$, giving high/low. For anchor $sft3$, the base direction is $-(u_1+u_2+u_3)$, and the two witness directions are $-(u_2+u_3)$ and $-u_3$, both sharing reverse chain components, giving high/high.

\section{Per-Family Single-Witness Geometry}
\label{app:family_geometry}
\begin{table}[t]
\centering
\small
\caption{\textbf{Per-family single-witness geometry.} Each row summarizes all LLM $\Nw{=}1$ parent-child-witness triplets in one family. Here $i$ is the evaluated child update, $j$ is the witness update, $\rho=C(p;i,j)$, and $\lambda=\|u_i\|/\|u_j\|$. The final column reports the median of the triplet-level ratio $\rho/\lambda$, which is the quantity compared with $1/2$ in \Cref{prop:rank_one}.}
\label{tab:family_rho_lambda}
\setlength{\tabcolsep}{4pt}
\begin{tabular}{lrrrr}
\toprule
Family & Triplets & median $\rho$ & median $\lambda$ & median $\rho/\lambda$ \\
\midrule
DeepSeek-R1-Distill-Qwen-1.5B & 72 & 0.0011 & 1.0000 & 4.86e-04 \\
Gemma-3-4B-it & 42 & 1.98e-04 & 1.0021 & 4.34e-04 \\
GPT-2 & 90 & 0.0087 & 1.0002 & 0.0056 \\
GPT-2 Medium & 90 & 0.0026 & 1.0000 & 0.0022 \\
GPT-2 XL & 56 & 6.44e-04 & 1.0015 & 3.15e-04 \\
Llama-3.2-1B-Instruct & 182 & 3.32e-04 & 1.0002 & 2.14e-04 \\
Meta-Llama-3-8B & 90 & 3.27e-04 & 1.0091 & 5.82e-04 \\
Mistral-7B-v0.3 & 90 & 6.37e-04 & 1.0005 & 8.50e-04 \\
Pythia-410M & 42 & 0.0081 & 1.0005 & 0.0106 \\
Qwen2.5-0.5B & 182 & 2.89e-04 & 1.0000 & 3.87e-04 \\
Qwen3-0.6B-Base & 72 & 4.14e-04 & 1.0041 & 7.96e-04 \\
Qwen3-8B & 42 & 0.0014 & 1.0000 & 0.0015 \\
SmolLM2-135M & 182 & 0.0114 & 1.0000 & 0.0139 \\
SmolLM2-1.7B-Instruct & 72 & 9.68e-06 & 1.0002 & 3.44e-06 \\
SmolLM2-360M-Instruct & 182 & 1.59e-04 & 1.0063 & 1.49e-05 \\
TinyLlama-1.1B-Chat-v1.0 & 56 & 1.03e-04 & 1.0040 & 1.49e-04 \\
\textbf{Overall} & \textbf{1542} & \textbf{6.37e-04} & \textbf{1.0000} & \textbf{8.08e-04} \\
\bottomrule
\end{tabular}
\end{table}

\FloatBarrier

\Cref{tab:family_rho_lambda} summarizes the geometry behind the single-witness direction test by family. The medians are computed over all LLM parent-child-witness triplets at $\Nw{=}1$, using the same $\rho$ and $\lambda$ notation as \Cref{sec:theory}. The ratio $\rho/\lambda$ is included because the rank-one condition can be written as $\rho/\lambda<1/2$. These summaries are descriptive and are separate from the failure-only list in \Cref{app:failure_cases}.

\section{VLM and Diffusion Witness Averaging}
\label{app:cross_domain_kwitness}
\begin{figure}[H]
    \centering
    \begin{subfigure}[t]{0.48\textwidth}\centering
        \includegraphics[width=\linewidth]{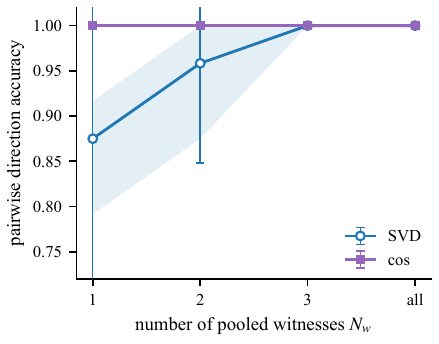}
        \caption{VLM}
        \label{fig:kwitness_sweep_vlm}
    \end{subfigure}\hfill
    \begin{subfigure}[t]{0.48\textwidth}\centering
        \includegraphics[width=\linewidth]{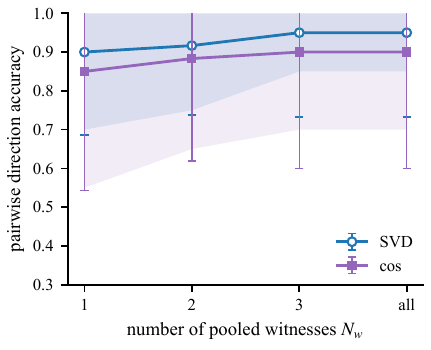}
        \caption{Diffusion}
        \label{fig:kwitness_sweep_sd}
    \end{subfigure}
    \caption{\textbf{Cross-domain $\Nw$-witness ablation.} Here $\Nw$ is the number of same-family witnesses averaged before making a decision. The main result uses $\Nw{=}1$; larger $\Nw$ tests whether pooling witnesses stabilizes the same directional signal on VLM and Diffusion families.}
    \label{fig:kwitness_sweep_crossdomain}
\end{figure}

We examine witness averaging on VLM and diffusion checkpoints. As shown in \Cref{fig:kwitness_sweep_crossdomain}, \textsc{Witness Overlap} already achieves high parent--child orientation accuracy with a single witness ($N_w=1$), and performance remains strong as more witnesses are averaged. SVD overlap shows a slight improvement with larger $N_w$, suggesting that pooling witnesses can reduce noise in the dominant update subspaces.

\section{Same-Family Model Merging}
\label{app:same_family_merges}

We test whether a merge of two descendants is still placed on the descendant side of their common root. We use two qualitatively different weight-space merge operators: TIES~\citep{yadav2023ties}, which combines root-relative task vectors while resolving conflicting updates, and SLERP~\citep{shoemake1985animating}, which interpolates directly between descendant weights. The labels $0.50/0.50$ and $0.25/0.75$ denote balanced and asymmetric nominal merge weights, respectively.

For root identification, we select four direct children from each family's full child set in a fixed registry order. Merging two of them leaves a four-checkpoint candidate set containing the root, the merged checkpoint, and two unchanged children. For direction, we sample ten child pairs from each complete family with $7$--$14$ children, merge each pair, and query the direction between the merged checkpoint and its root. We evaluate each root--merge direction using either one eligible witness ($\Nw{=}1$) or the average over all eligible witnesses.

\begin{table}[H]
\centering
\footnotesize
\caption{\textbf{Same-family merge evaluation.} We merge two descendants that share a root using TIES or SLERP. Root ID reports accuracy over the resulting four-checkpoint candidate sets. Direction reports the accuracy of placing the merged checkpoint on the descendant side of the root using one or all eligible witnesses. Each entry gives Frobenius cosine / SVD top-$k{=}16$.}
\label{tab:same_family_merges}
\setlength{\tabcolsep}{4pt}
\begin{tabular}{lccc}
\toprule
Merge & Root ID & $\Nw{=}1$ direction & All-witness direction \\
\midrule
TIES $0.50/0.50$  & $93.75\% / 87.50\%$ & $97.03\% / 96.95\%$ & $100.00\% / 100.00\%$ \\
TIES $0.25/0.75$  & $93.75\% / 93.75\%$ & $97.03\% / 96.88\%$ & $100.00\% / 100.00\%$ \\
SLERP $0.50/0.50$ & $93.75\% / 87.50\%$ & $96.09\% / 96.41\%$ & $97.50\% / 100.00\%$ \\
SLERP $0.25/0.75$ & $81.25\% / 87.50\%$ & $95.70\% / 96.56\%$ & $98.13\% / 100.00\%$ \\
\bottomrule
\end{tabular}
\end{table}

The merged checkpoint is almost always placed on the descendant side. With one witness, direction accuracy is $96.88\%$--$97.03\%$ for TIES and $95.70\%$--$96.56\%$ for SLERP. Using all witnesses, both TIES metrics and SLERP with SVD reach $100\%$; SLERP with Frobenius cosine reaches $97.50\%$ for the balanced merge and $98.13\%$ for the asymmetric merge. This experiment addresses root-relative placement only: it neither identifies the contributor set nor orients individual contributor-to-merge edges.

\FloatBarrier

\section{Quantization Robustness Test}
\label{app:int8_quantization}

We evaluate symmetric per-tensor signed INT8 quantize--dequantize in two settings. First, we apply the same rule to every checkpoint in three complete LLM families. The quantized checkpoints are recovered into a common dense parameterization before computing weight differences.

\begin{table}[H]
\centering
\footnotesize
\caption{\textbf{All-checkpoint INT8 quantization.} The same symmetric per-tensor signed INT8 quantize--dequantize rule is applied to every checkpoint in each family. Direction is evaluated with one or all eligible witnesses.}
\label{tab:quantization_all_models}
\setlength{\tabcolsep}{4pt}
\begin{tabular}{llccc}
\toprule
Family & Metric & Root ID & $\Nw{=}1$ direction & All-witness direction \\
\midrule
Llama-3.2-1B-Instruct & Cosine & $1/1$ & $182/182$ & $14/14$ \\
 & SVD & $1/1$ & $179/182$ & $14/14$ \\
Qwen2.5-0.5B & Cosine & $1/1$ & $169/182$ & $14/14$ \\
 & SVD & $1/1$ & $159/182$ & $14/14$ \\
Qwen3-0.6B-Base & Cosine & $1/1$ & $72/72$ & $9/9$ \\
 & SVD & $1/1$ & $68/72$ & $9/9$ \\
\midrule
\textbf{Overall} & \textbf{Cosine} & $\mathbf{3/3}$ & $\mathbf{423/436\ (97.0\%)}$ & $\mathbf{37/37}$ \\
 & \textbf{SVD} & $\mathbf{3/3}$ & $\mathbf{406/436\ (93.1\%)}$ & $\mathbf{37/37}$ \\
\bottomrule
\end{tabular}
\end{table}

INT8 preserves all three root predictions and all $37$ all-witness direction decisions. With one witness, cosine reaches $423/436$ ($97.0\%$) and SVD reaches $406/436$ ($93.1\%$).

Second, we evaluate a representation-mismatch setting across five complete LLM families containing $70$ checkpoints and $65$ parent--child edges. For each triplet, the parent and witness remain clean while only the queried child is quantized. This produces $800$ single-witness decisions per metric.

\begin{table}[H]
\centering
\footnotesize
\caption{\textbf{Child-only INT8 quantization.} The parent and witness remain clean while only the queried child is quantized. The clean control uses the same five families and triplets.}
\label{tab:quantization_child_only}
\setlength{\tabcolsep}{6pt}
\begin{tabular}{lcc}
\toprule
Setting & Cosine, single / all witnesses & SVD, single / all witnesses \\
\midrule
Clean control & $752/800\ (94.00\%)$ / $64/65$ & $731/800\ (91.38\%)$ / $65/65$ \\
Child-only INT8 & $794/800\ (99.25\%)$ / $65/65$ & $794/800\ (99.25\%)$ / $65/65$ \\
\bottomrule
\end{tabular}
\end{table}

No decision that was correct on the clean checkpoints becomes incorrect after child-only INT8. The perturbation instead corrects $42$ cosine decisions and $63$ SVD decisions that were previously wrong. This apparent improvement does not mean that quantization generally improves provenance inference. The child-only quantization residual is shared by both difference vectors when the child is the anchor, but enters only one difference vector when the parent is the anchor. This strengthens the overlap asymmetry in this specific setting.

\FloatBarrier

\section{Anchor-Role Asymmetry}
\label{app:anchor_role_asymmetry}
\FloatBarrier
\begin{table}[H]
\centering
\scriptsize
\caption{\textbf{Overlap distribution by anchor role.} For each parent--child pair we average the witness overlap over all in-family witnesses, separately for the two anchor choices (parent or child). The table aggregates these per-pair values across the LLM, VLM, and Diffusion evaluations. Both statistics show the same asymmetry: parent-anchor overlap is usually smaller than child-anchor overlap.}
\label{tab:anchor_overlap}
\setlength{\tabcolsep}{3pt}
\resizebox{\textwidth}{!}{%
\begin{tabular}{ll ccc ccc ccc}
\toprule
Method & Anchor role & \multicolumn{3}{c}{LLM} & \multicolumn{3}{c}{VLM} & \multicolumn{3}{c}{Diffusion} \\
\cmidrule(lr){3-5}\cmidrule(lr){6-8}\cmidrule(lr){9-11}
& & Mean & Median & Std & Mean & Median & Std & Mean & Median & Std \\
\midrule
Frobenius cosine & parent & 0.0232 & 0.0025 & 0.0482 & 0.0350 & 0.0009 & 0.0546 & 0.0950 & 0.0294 & 0.1615 \\
Frobenius cosine & child & 0.5527 & 0.5678 & 0.2704 & 0.5140 & 0.4647 & 0.3056 & 0.4278 & 0.3845 & 0.2582 \\
\addlinespace[1pt]
SVD top-$k{=}16$ & parent & 0.0725 & 0.0405 & 0.0867 & 0.1034 & 0.0619 & 0.0920 & 0.2454 & 0.1543 & 0.2216 \\
SVD top-$k{=}16$ & child & 0.3961 & 0.3950 & 0.1720 & 0.4181 & 0.4629 & 0.2142 & 0.4537 & 0.4326 & 0.2338 \\
\bottomrule
\end{tabular}
}
\end{table}

\FloatBarrier
\section{Single-Witness Failure Cases}
\label{app:failure_cases}

\Cref{tab:cos_failure_cases} lists the $73$ LLM triplets that are misoriented by the Frobenius cosine at $\Nw{=}1$. We use the notation from \Cref{sec:theory}: $i$ is the evaluated child update, $j$ is the witness update, $\rho=C(p;i,j)$, and $\eta=\|u_j\|/\|u_i\|=1/\lambda$.

\begingroup
\tiny
\setlength{\tabcolsep}{2pt}
\begin{longtable}{@{}p{0.39\textwidth}p{0.39\textwidth}rrp{0.08\textwidth}@{}}
\caption{\textbf{Frobenius-cosine single-witness failures.} Each row is one misoriented LLM triplet at $\Nw{=}1$. Here $i$ is the child update, $j$ is the witness update, $\rho=C(p;i,j)$, and $\eta=\|u_j\|/\|u_i\|=1/\lambda$. The high-$\rho$ and high-$\eta$ tags mark the top $10\%$ of all cosine triplets, with thresholds $\rho\ge 0.0247$ and $\eta\ge 45.16$.}
\label{tab:cos_failure_cases}\\
\toprule
Child repository & Witness repository & $\rho$ & $\eta$ & Tag \\
\midrule
\endfirsthead
\caption[]{\textbf{Frobenius-cosine single-witness failures} continued.}\\
\toprule
Child repository & Witness repository & $\rho$ & $\eta$ & Tag \\
\midrule
\endhead
\midrule
\multicolumn{5}{r}{Continued on next page}\\
\endfoot
\bottomrule
\endlastfoot
\texttt{INSAIT-\allowbreak{}Institute/\allowbreak{}BgGPT-\allowbreak{}Gemma-\allowbreak{}3-\allowbreak{}4B-\allowbreak{}IT} & \texttt{INSAIT-\allowbreak{}Institute/\allowbreak{}MamayLM-\allowbreak{}Gemma-\allowbreak{}3-\allowbreak{}4B-\allowbreak{}IT-\allowbreak{}v1.\allowbreak{}0} & 0.5841 & 0.89 & high-$\rho$ \\
\texttt{INSAIT-\allowbreak{}Institute/\allowbreak{}MamayLM-\allowbreak{}Gemma-\allowbreak{}3-\allowbreak{}4B-\allowbreak{}IT-\allowbreak{}v1.\allowbreak{}0} & \texttt{INSAIT-\allowbreak{}Institute/\allowbreak{}BgGPT-\allowbreak{}Gemma-\allowbreak{}3-\allowbreak{}4B-\allowbreak{}IT} & 0.5841 & 1.12 & high-$\rho$ \\
\texttt{Ssarion/\allowbreak{}gpt2-\allowbreak{}multi-\allowbreak{}news} & \texttt{bipin/\allowbreak{}malayalam-\allowbreak{}gpt2} & 0.0319 & 97.36 & both \\
\texttt{Ssarion/\allowbreak{}gpt2-\allowbreak{}multi-\allowbreak{}news} & \texttt{jackoyoungblood/\allowbreak{}TinyStoriesTest} & 0.0325 & 95.51 & both \\
\texttt{bipin/\allowbreak{}malayalam-\allowbreak{}gpt2} & \texttt{jackoyoungblood/\allowbreak{}TinyStoriesTest} & 0.9648 & 0.98 & high-$\rho$ \\
\texttt{crumb/\allowbreak{}qrstudy-\allowbreak{}gpt2-\allowbreak{}2-\allowbreak{}4} & \texttt{bipin/\allowbreak{}malayalam-\allowbreak{}gpt2} & 0.0028 & 405.80 & high-$\eta$ \\
\texttt{crumb/\allowbreak{}qrstudy-\allowbreak{}gpt2-\allowbreak{}2-\allowbreak{}4} & \texttt{jackoyoungblood/\allowbreak{}TinyStoriesTest} & 0.0028 & 398.09 & high-$\eta$ \\
\texttt{jackoyoungblood/\allowbreak{}TinyStoriesTest} & \texttt{bipin/\allowbreak{}malayalam-\allowbreak{}gpt2} & 0.9648 & 1.02 & high-$\rho$ \\
\texttt{juancopi81/\allowbreak{}gpt2-\allowbreak{}finetuned-\allowbreak{}yannic-\allowbreak{}test} & \texttt{bipin/\allowbreak{}malayalam-\allowbreak{}gpt2} & 0.0382 & 81.50 & both \\
\texttt{juancopi81/\allowbreak{}gpt2-\allowbreak{}finetuned-\allowbreak{}yannic-\allowbreak{}test} & \texttt{jackoyoungblood/\allowbreak{}TinyStoriesTest} & 0.0389 & 79.95 & both \\
\texttt{juancopi81/\allowbreak{}gpt2-\allowbreak{}finetuned-\allowbreak{}yannic-\allowbreak{}test} & \texttt{kpriyanshu256/\allowbreak{}gpt-\allowbreak{}ya2-\allowbreak{}v2} & 0.1261 & 4.17 & high-$\rho$ \\
\texttt{nicholasKluge/\allowbreak{}Aira-\allowbreak{}2-\allowbreak{}124M} & \texttt{bipin/\allowbreak{}malayalam-\allowbreak{}gpt2} & 0.1140 & 27.47 & high-$\rho$ \\
\texttt{nicholasKluge/\allowbreak{}Aira-\allowbreak{}2-\allowbreak{}124M} & \texttt{jackoyoungblood/\allowbreak{}TinyStoriesTest} & 0.1165 & 26.95 & high-$\rho$ \\
\texttt{masani/\allowbreak{}sft\_\allowbreak{}checkpoints} & \texttt{masani/\allowbreak{}2025-\allowbreak{}04-\allowbreak{}02\_\allowbreak{}14-\allowbreak{}52-\allowbreak{}39} & 0.0178 & 32.73 & neither \\
\texttt{OpenRLHF/\allowbreak{}Llama-\allowbreak{}3-\allowbreak{}8b-\allowbreak{}sft-\allowbreak{}mixture} & \texttt{openchat/\allowbreak{}openchat-\allowbreak{}3.\allowbreak{}6-\allowbreak{}8b-\allowbreak{}20240522} & 0.1736 & 3.24 & high-$\rho$ \\
\texttt{Weyaxi/\allowbreak{}Einstein-\allowbreak{}v6.\allowbreak{}1-\allowbreak{}Llama3-\allowbreak{}8B} & \texttt{OpenRLHF/\allowbreak{}Llama-\allowbreak{}3-\allowbreak{}8b-\allowbreak{}sft-\allowbreak{}mixture} & 0.3928 & 1.44 & high-$\rho$ \\
\texttt{Weyaxi/\allowbreak{}Einstein-\allowbreak{}v6.\allowbreak{}1-\allowbreak{}Llama3-\allowbreak{}8B} & \texttt{openchat/\allowbreak{}openchat-\allowbreak{}3.\allowbreak{}6-\allowbreak{}8b-\allowbreak{}20240522} & 0.1454 & 4.65 & high-$\rho$ \\
\texttt{lomahony/\allowbreak{}eleuther-\allowbreak{}pythia410m-\allowbreak{}hh-\allowbreak{}dpo} & \texttt{kykim0/\allowbreak{}pythia-\allowbreak{}410m-\allowbreak{}tulu-\allowbreak{}v2-\allowbreak{}mix} & 0.0137 & 45.23 & high-$\eta$ \\
\texttt{lomahony/\allowbreak{}eleuther-\allowbreak{}pythia410m-\allowbreak{}hh-\allowbreak{}dpo} & \texttt{lomahony/\allowbreak{}pythia-\allowbreak{}410m-\allowbreak{}helpful-\allowbreak{}dpo} & 0.5295 & 0.97 & high-$\rho$ \\
\texttt{lomahony/\allowbreak{}pythia-\allowbreak{}410m-\allowbreak{}helpful-\allowbreak{}dpo} & \texttt{kykim0/\allowbreak{}pythia-\allowbreak{}410m-\allowbreak{}tulu-\allowbreak{}v2-\allowbreak{}mix} & 0.0152 & 46.69 & high-$\eta$ \\
\texttt{lomahony/\allowbreak{}pythia-\allowbreak{}410m-\allowbreak{}helpful-\allowbreak{}dpo} & \texttt{lomahony/\allowbreak{}eleuther-\allowbreak{}pythia410m-\allowbreak{}hh-\allowbreak{}dpo} & 0.5295 & 1.03 & high-$\rho$ \\
\texttt{lomahony/\allowbreak{}pythia-\allowbreak{}410m-\allowbreak{}helpful-\allowbreak{}dpo} & \texttt{nnheui/\allowbreak{}pythia-\allowbreak{}410m-\allowbreak{}sft-\allowbreak{}full} & 0.0535 & 10.16 & high-$\rho$ \\
\texttt{BounharAbdelaziz/\allowbreak{}Qwen2.\allowbreak{}5-\allowbreak{}0.\allowbreak{}5B-\allowbreak{}DPO-\allowbreak{}English-\allowbreak{}Orca} & \texttt{Saralatifi/\allowbreak{}Qwen2.\allowbreak{}5-\allowbreak{}0.\allowbreak{}5B-\allowbreak{}DPO} & 1.0000 & 1.00 & high-$\rho$ \\
\texttt{BounharAbdelaziz/\allowbreak{}Qwen2.\allowbreak{}5-\allowbreak{}0.\allowbreak{}5B-\allowbreak{}DPO-\allowbreak{}English-\allowbreak{}Orca} & \texttt{shawhin/\allowbreak{}Qwen2.\allowbreak{}5-\allowbreak{}0.\allowbreak{}5B-\allowbreak{}DPO} & 0.9999 & 1.00 & high-$\rho$ \\
\texttt{Saralatifi/\allowbreak{}Qwen2.\allowbreak{}5-\allowbreak{}0.\allowbreak{}5B-\allowbreak{}DPO} & \texttt{BounharAbdelaziz/\allowbreak{}Qwen2.\allowbreak{}5-\allowbreak{}0.\allowbreak{}5B-\allowbreak{}DPO-\allowbreak{}English-\allowbreak{}Orca} & 1.0000 & 1.00 & high-$\rho$ \\
\texttt{Saralatifi/\allowbreak{}Qwen2.\allowbreak{}5-\allowbreak{}0.\allowbreak{}5B-\allowbreak{}DPO} & \texttt{shawhin/\allowbreak{}Qwen2.\allowbreak{}5-\allowbreak{}0.\allowbreak{}5B-\allowbreak{}DPO} & 1.0000 & 1.00 & high-$\rho$ \\
\texttt{shawhin/\allowbreak{}Qwen2.\allowbreak{}5-\allowbreak{}0.\allowbreak{}5B-\allowbreak{}DPO} & \texttt{BounharAbdelaziz/\allowbreak{}Qwen2.\allowbreak{}5-\allowbreak{}0.\allowbreak{}5B-\allowbreak{}DPO-\allowbreak{}English-\allowbreak{}Orca} & 0.9999 & 1.00 & high-$\rho$ \\
\texttt{shawhin/\allowbreak{}Qwen2.\allowbreak{}5-\allowbreak{}0.\allowbreak{}5B-\allowbreak{}DPO} & \texttt{Saralatifi/\allowbreak{}Qwen2.\allowbreak{}5-\allowbreak{}0.\allowbreak{}5B-\allowbreak{}DPO} & 1.0000 & 1.00 & high-$\rho$ \\
\texttt{open-\allowbreak{}thoughts/\allowbreak{}OpenThinker-\allowbreak{}Agent-\allowbreak{}v1} & \texttt{open-\allowbreak{}thoughts/\allowbreak{}OpenThinker-\allowbreak{}Agent-\allowbreak{}v1-\allowbreak{}SFT} & 0.9985 & 1.00 & high-$\rho$ \\
\texttt{open-\allowbreak{}thoughts/\allowbreak{}OpenThinker-\allowbreak{}Agent-\allowbreak{}v1-\allowbreak{}SFT} & \texttt{open-\allowbreak{}thoughts/\allowbreak{}OpenThinker-\allowbreak{}Agent-\allowbreak{}v1} & 0.9985 & 1.00 & high-$\rho$ \\
\texttt{MUTSC/\allowbreak{}SmolLM2-\allowbreak{}FT-\allowbreak{}ORPO} & \texttt{HuggingFaceTB/\allowbreak{}smollm2-\allowbreak{}135M-\allowbreak{}SFT-\allowbreak{}Only} & 0.0798 & 16.05 & high-$\rho$ \\
\texttt{MUTSC/\allowbreak{}SmolLM2-\allowbreak{}FT-\allowbreak{}ORPO} & \texttt{hsila/\allowbreak{}SmolLM2-\allowbreak{}135M-\allowbreak{}ORPO} & 0.9634 & 1.01 & high-$\rho$ \\
\texttt{Nels2/\allowbreak{}SmolLM2-\allowbreak{}FT-\allowbreak{}SQL-\allowbreak{}Context} & \texttt{ParitKansal/\allowbreak{}SmolLM2-\allowbreak{}135M-\allowbreak{}SFT-\allowbreak{}smoltalk} & 0.9490 & 0.98 & high-$\rho$ \\
\texttt{Nels2/\allowbreak{}SmolLM2-\allowbreak{}FT-\allowbreak{}SQL-\allowbreak{}Context} & \texttt{puettmann/\allowbreak{}SmolLM2-\allowbreak{}135M-\allowbreak{}Instruct-\allowbreak{}Smol-\allowbreak{}Course} & 0.9899 & 1.00 & high-$\rho$ \\
\texttt{ParitKansal/\allowbreak{}SmolLM2-\allowbreak{}135M-\allowbreak{}SFT-\allowbreak{}smoltalk} & \texttt{Nels2/\allowbreak{}SmolLM2-\allowbreak{}FT-\allowbreak{}SQL-\allowbreak{}Context} & 0.9490 & 1.02 & high-$\rho$ \\
\texttt{ParitKansal/\allowbreak{}SmolLM2-\allowbreak{}135M-\allowbreak{}SFT-\allowbreak{}smoltalk} & \texttt{puettmann/\allowbreak{}SmolLM2-\allowbreak{}135M-\allowbreak{}Instruct-\allowbreak{}Smol-\allowbreak{}Course} & 0.9445 & 1.02 & high-$\rho$ \\
\texttt{gotoplanb/\allowbreak{}SmolLM2-\allowbreak{}FT-\allowbreak{}MyDataset} & \texttt{HuggingFaceTB/\allowbreak{}smollm2-\allowbreak{}135M-\allowbreak{}SFT-\allowbreak{}Only} & 0.0256 & 22.21 & high-$\rho$ \\
\texttt{gotoplanb/\allowbreak{}SmolLM2-\allowbreak{}FT-\allowbreak{}MyDataset} & \texttt{Nels2/\allowbreak{}SmolLM2-\allowbreak{}FT-\allowbreak{}SQL-\allowbreak{}Context} & 0.7200 & 1.63 & high-$\rho$ \\
\texttt{gotoplanb/\allowbreak{}SmolLM2-\allowbreak{}FT-\allowbreak{}MyDataset} & \texttt{ParitKansal/\allowbreak{}SmolLM2-\allowbreak{}135M-\allowbreak{}SFT-\allowbreak{}smoltalk} & 0.6311 & 1.59 & high-$\rho$ \\
\texttt{gotoplanb/\allowbreak{}SmolLM2-\allowbreak{}FT-\allowbreak{}MyDataset} & \texttt{puettmann/\allowbreak{}SmolLM2-\allowbreak{}135M-\allowbreak{}Instruct-\allowbreak{}Smol-\allowbreak{}Course} & 0.7120 & 1.63 & high-$\rho$ \\
\texttt{hsila/\allowbreak{}SmolLM2-\allowbreak{}135M-\allowbreak{}ORPO} & \texttt{HuggingFaceTB/\allowbreak{}smollm2-\allowbreak{}135M-\allowbreak{}SFT-\allowbreak{}Only} & 0.0731 & 15.85 & high-$\rho$ \\
\texttt{hsila/\allowbreak{}SmolLM2-\allowbreak{}135M-\allowbreak{}ORPO} & \texttt{MUTSC/\allowbreak{}SmolLM2-\allowbreak{}FT-\allowbreak{}ORPO} & 0.9634 & 0.99 & high-$\rho$ \\
\texttt{lhoestq/\allowbreak{}finetune\_\allowbreak{}smollm2\_\allowbreak{}python} & \texttt{Ellight/\allowbreak{}SmolLM2-\allowbreak{}135M-\allowbreak{}text-\allowbreak{}to-\allowbreak{}sql} & 0.0004 & 99531.4 & high-$\eta$ \\
\texttt{lhoestq/\allowbreak{}finetune\_\allowbreak{}smollm2\_\allowbreak{}python} & \texttt{Ellight/\allowbreak{}code-\allowbreak{}smolLM2-\allowbreak{}135m-\allowbreak{}text-\allowbreak{}to-\allowbreak{}sql} & 0.0003 & 114216.7 & high-$\eta$ \\
\texttt{lhoestq/\allowbreak{}finetune\_\allowbreak{}smollm2\_\allowbreak{}python} & \texttt{MUTSC/\allowbreak{}SmolLM2-\allowbreak{}FT-\allowbreak{}ORPO} & 0.0002 & 43371.1 & high-$\eta$ \\
\texttt{lhoestq/\allowbreak{}finetune\_\allowbreak{}smollm2\_\allowbreak{}python} & \texttt{Nels2/\allowbreak{}SmolLM2-\allowbreak{}FT-\allowbreak{}SQL-\allowbreak{}Context} & 0.0003 & 51038.3 & high-$\eta$ \\
\texttt{lhoestq/\allowbreak{}finetune\_\allowbreak{}smollm2\_\allowbreak{}python} & \texttt{ParitKansal/\allowbreak{}SmolLM2-\allowbreak{}135M-\allowbreak{}SFT-\allowbreak{}smoltalk} & 0.0003 & 49809.4 & high-$\eta$ \\
\texttt{lhoestq/\allowbreak{}finetune\_\allowbreak{}smollm2\_\allowbreak{}python} & \texttt{frankenstein-\allowbreak{}ai/\allowbreak{}admin-\allowbreak{}SmolLM2-\allowbreak{}135M-\allowbreak{}20251204\_\allowbreak{}194139} & 0.0004 & 8813.4 & high-$\eta$ \\
\texttt{lhoestq/\allowbreak{}finetune\_\allowbreak{}smollm2\_\allowbreak{}python} & \texttt{gotoplanb/\allowbreak{}SmolLM2-\allowbreak{}FT-\allowbreak{}MyDataset} & 0.0003 & 31339.6 & high-$\eta$ \\
\texttt{lhoestq/\allowbreak{}finetune\_\allowbreak{}smollm2\_\allowbreak{}python} & \texttt{hsila/\allowbreak{}SmolLM2-\allowbreak{}135M-\allowbreak{}ORPO} & 0.0003 & 43922.9 & high-$\eta$ \\
\texttt{lhoestq/\allowbreak{}finetune\_\allowbreak{}smollm2\_\allowbreak{}python} & \texttt{mnoukhov/\allowbreak{}SmolLM2-\allowbreak{}135M-\allowbreak{}tldr-\allowbreak{}sft} & 0.0001 & 26928.5 & high-$\eta$ \\
\texttt{lhoestq/\allowbreak{}finetune\_\allowbreak{}smollm2\_\allowbreak{}python} & \texttt{puettmann/\allowbreak{}SmolLM2-\allowbreak{}135M-\allowbreak{}Instruct-\allowbreak{}Smol-\allowbreak{}Course} & 0.0003 & 50971.1 & high-$\eta$ \\
\texttt{peaceAsh/\allowbreak{}smolcourse\_\allowbreak{}chapter2\_\allowbreak{}ORPO} & \texttt{Ashed00/\allowbreak{}SmolMath-\allowbreak{}135M} & 0.0024 & 400.44 & high-$\eta$ \\
\texttt{peaceAsh/\allowbreak{}smolcourse\_\allowbreak{}chapter2\_\allowbreak{}ORPO} & \texttt{Ellight/\allowbreak{}SmolLM2-\allowbreak{}135M-\allowbreak{}text-\allowbreak{}to-\allowbreak{}sql} & 0.0175 & 164.27 & high-$\eta$ \\
\texttt{peaceAsh/\allowbreak{}smolcourse\_\allowbreak{}chapter2\_\allowbreak{}ORPO} & \texttt{Ellight/\allowbreak{}code-\allowbreak{}smolLM2-\allowbreak{}135m-\allowbreak{}text-\allowbreak{}to-\allowbreak{}sql} & 0.0218 & 188.50 & high-$\eta$ \\
\texttt{peaceAsh/\allowbreak{}smolcourse\_\allowbreak{}chapter2\_\allowbreak{}ORPO} & \texttt{HuggingFaceTB/\allowbreak{}smollm2-\allowbreak{}135M-\allowbreak{}SFT-\allowbreak{}Only} & 0.0159 & 1148.9 & high-$\eta$ \\
\texttt{peaceAsh/\allowbreak{}smolcourse\_\allowbreak{}chapter2\_\allowbreak{}ORPO} & \texttt{MUTSC/\allowbreak{}SmolLM2-\allowbreak{}FT-\allowbreak{}ORPO} & 0.1532 & 71.58 & both \\
\texttt{peaceAsh/\allowbreak{}smolcourse\_\allowbreak{}chapter2\_\allowbreak{}ORPO} & \texttt{Nels2/\allowbreak{}SmolLM2-\allowbreak{}FT-\allowbreak{}SQL-\allowbreak{}Context} & 0.0602 & 84.23 & both \\
\texttt{peaceAsh/\allowbreak{}smolcourse\_\allowbreak{}chapter2\_\allowbreak{}ORPO} & \texttt{ParitKansal/\allowbreak{}SmolLM2-\allowbreak{}135M-\allowbreak{}SFT-\allowbreak{}smoltalk} & 0.0669 & 82.21 & both \\
\texttt{peaceAsh/\allowbreak{}smolcourse\_\allowbreak{}chapter2\_\allowbreak{}ORPO} & \texttt{gotoplanb/\allowbreak{}SmolLM2-\allowbreak{}FT-\allowbreak{}MyDataset} & 0.0915 & 51.72 & both \\
\texttt{peaceAsh/\allowbreak{}smolcourse\_\allowbreak{}chapter2\_\allowbreak{}ORPO} & \texttt{hsila/\allowbreak{}SmolLM2-\allowbreak{}135M-\allowbreak{}ORPO} & 0.1519 & 72.49 & both \\
\texttt{peaceAsh/\allowbreak{}smolcourse\_\allowbreak{}chapter2\_\allowbreak{}ORPO} & \texttt{puettmann/\allowbreak{}SmolLM2-\allowbreak{}135M-\allowbreak{}Instruct-\allowbreak{}Smol-\allowbreak{}Course} & 0.0608 & 84.12 & both \\
\texttt{puettmann/\allowbreak{}SmolLM2-\allowbreak{}135M-\allowbreak{}Instruct-\allowbreak{}Smol-\allowbreak{}Course} & \texttt{Nels2/\allowbreak{}SmolLM2-\allowbreak{}FT-\allowbreak{}SQL-\allowbreak{}Context} & 0.9899 & 1.00 & high-$\rho$ \\
\texttt{puettmann/\allowbreak{}SmolLM2-\allowbreak{}135M-\allowbreak{}Instruct-\allowbreak{}Smol-\allowbreak{}Course} & \texttt{ParitKansal/\allowbreak{}SmolLM2-\allowbreak{}135M-\allowbreak{}SFT-\allowbreak{}smoltalk} & 0.9445 & 0.98 & high-$\rho$ \\
\texttt{Michaelj1/\allowbreak{}INSTRUCT\_\allowbreak{}smolLM2-\allowbreak{}360M-\allowbreak{}finetuned-\allowbreak{}wikitext2-\allowbreak{}raw-\allowbreak{}v1} & \texttt{prithivMLmods/\allowbreak{}SmolLM2-\allowbreak{}CoT-\allowbreak{}360M} & 0.0127 & 43.73 & neither \\
\texttt{TheBlueObserver/\allowbreak{}SmolLM2-\allowbreak{}360M-\allowbreak{}Instruct-\allowbreak{}MLX} & \texttt{Michaelj1/\allowbreak{}INSTRUCT\_\allowbreak{}smolLM2-\allowbreak{}360M-\allowbreak{}finetuned-\allowbreak{}wikitext2-\allowbreak{}raw-\allowbreak{}v1} & 0.0001 & 2.45e+06 & high-$\eta$ \\
\texttt{TheBlueObserver/\allowbreak{}SmolLM2-\allowbreak{}360M-\allowbreak{}Instruct-\allowbreak{}MLX} & \texttt{ThatsGroes/\allowbreak{}SmolLM2-\allowbreak{}360M-\allowbreak{}Instruct-\allowbreak{}summarizer} & 0.0002 & 5.16e+08 & high-$\eta$ \\
\texttt{TheBlueObserver/\allowbreak{}SmolLM2-\allowbreak{}360M-\allowbreak{}Instruct-\allowbreak{}MLX} & \texttt{ericlewis/\allowbreak{}infinite-\allowbreak{}craft-\allowbreak{}smollm2-\allowbreak{}360m-\allowbreak{}full-\allowbreak{}grpo} & 0.0001& 137562.1 & high-$\eta$ \\
\texttt{TheBlueObserver/\allowbreak{}SmolLM2-\allowbreak{}360M-\allowbreak{}Instruct-\allowbreak{}MLX} & \texttt{motexture/\allowbreak{}SmolLCoder-\allowbreak{}360M-\allowbreak{}Instruct} & 0.0001 & 2.81e+06 & high-$\eta$ \\
\texttt{TheBlueObserver/\allowbreak{}SmolLM2-\allowbreak{}360M-\allowbreak{}Instruct-\allowbreak{}MLX} & \texttt{prithivMLmods/\allowbreak{}SmolLM2-\allowbreak{}CoT-\allowbreak{}360M} & 0.0001 & 1.07e+08 & high-$\eta$ \\
\texttt{TheBlueObserver/\allowbreak{}SmolLM2-\allowbreak{}360M-\allowbreak{}Instruct-\allowbreak{}MLX} & \texttt{thatupiso/\allowbreak{}SmolLM2-\allowbreak{}360M-\allowbreak{}Instruct-\allowbreak{}K12-\allowbreak{}5000} & 0.0001 & 3.60e+06 & high-$\eta$ \\
\texttt{h4rz3rk4s3/\allowbreak{}TinyNewsLlama-\allowbreak{}1.\allowbreak{}1B} & \texttt{h4rz3rk4s3/\allowbreak{}TinyParlaMintLlama-\allowbreak{}1.\allowbreak{}1B} & 0.5607 & 1.09 & high-$\rho$ \\
\texttt{h4rz3rk4s3/\allowbreak{}TinyParlaMintLlama-\allowbreak{}1.\allowbreak{}1B} & \texttt{h4rz3rk4s3/\allowbreak{}TinyNewsLlama-\allowbreak{}1.\allowbreak{}1B} & 0.5607 & 0.91 & high-$\rho$ \\
\end{longtable}
\endgroup

\section{Pruning Stress Test}
\label{app:magnitude_pruning_stress}

We evaluate shape-preserving magnitude pruning as a stress test on the dense two-dimensional projection tensors used by our scores. At sparsity $s$, the smallest $s$ fraction of entries by absolute value is set to zero, while tensor names and shapes are unchanged. We consider three settings. In the child-only setting, only the queried child checkpoint is pruned, and the base and witness checkpoints remain clean. In the half-descendant setting, a fixed seeded random half of the GPT-2 descendants is pruned, so a triplet uses a pruned endpoint whenever its child or witness belongs to this subset. In the all-descendant setting, every GPT-2 descendant is pruned while the base checkpoint remains clean. The full pruning results for the three settings are reported in ~\Cref{tab:pruning_nw_all_descendants,tab:pruning_nw_child_only,tab:pruning_nw_half_descendants}. On GPT-2, child-only pruning is stable through 50\% sparsity: at 50\%, SVD and Frobenius cosine both achieve $80.0\%,97.5\%,100.0\%,100.0\%$ direction accuracy for $\Nw=1,2,3,\mathrm{all}$. The half-descendant and all-descendant settings are harder because descendant endpoints are sparsified. In the all-descendant setting, SVD is more robust than Frobenius cosine at moderate sparsity: at 10\% pruning, SVD achieves $77.8\%,76.1\%,74.3\%,70.0\%$, while Frobenius cosine drops to $53.3\%,52.5\%,49.5\%,50.0\%$. Averaging across more witnesses improves accuracy and reduces the side effects of pruning in milder settings, but pruning all descendants is a stronger perturbation that degrades both scores at high sparsity.

\begin{table}[H]
\centering
\tiny
\caption{\textbf{Child-only pruning stress test.} Only the evaluated child checkpoint is pruned, the parent and witness endpoints remain clean. Entries are parent-child direction accuracy after aggregating exactly $\Nw$ same-family witnesses, with $\Nw{=}\mathrm{all}$ using all eligible witnesses.}
\label{tab:pruning_nw_child_only}
\setlength{\tabcolsep}{2pt}
\renewcommand{\arraystretch}{0.92}
\resizebox{0.82\textwidth}{!}{%
\begin{tabular}{lllrrrr}
\toprule
Model & Rule & Magnitude pruning & $\Nw{=}1$ & $\Nw{=}2$ & $\Nw{=}3$ & $\Nw{=}\mathrm{all}$ \\
\midrule
GPT-2 & Frobenius cosine & 0\% & 87.8\% & 98.1\% & 99.8\% & 100.0\% \\
GPT-2 & Frobenius cosine & 2\% & 87.8\% & 98.1\% & 99.8\% & 100.0\% \\
GPT-2 & Frobenius cosine & 5\% & 86.7\% & 97.8\% & 99.9\% & 100.0\% \\
GPT-2 & Frobenius cosine & 10\% & 84.4\% & 98.1\% & 100.0\% & 100.0\% \\
GPT-2 & Frobenius cosine & 30\% & 80.0\% & 97.5\% & 100.0\% & 100.0\% \\
GPT-2 & Frobenius cosine & 50\% & 80.0\% & 97.5\% & 100.0\% & 100.0\% \\
\addlinespace[1pt]
GPT-2 & SVD top-$k{=}16$ & 0\% & 88.9\% & 96.7\% & 99.2\% & 100.0\% \\
GPT-2 & SVD top-$k{=}16$ & 2\% & 93.3\% & 97.5\% & 99.2\% & 100.0\% \\
GPT-2 & SVD top-$k{=}16$ & 5\% & 90.0\% & 97.2\% & 99.2\% & 100.0\% \\
GPT-2 & SVD top-$k{=}16$ & 10\% & 91.1\% & 98.1\% & 99.3\% & 100.0\% \\
GPT-2 & SVD top-$k{=}16$ & 30\% & 80.0\% & 97.5\% & 100.0\% & 100.0\% \\
GPT-2 & SVD top-$k{=}16$ & 50\% & 80.0\% & 97.5\% & 100.0\% & 100.0\% \\
\bottomrule
\end{tabular}
}
\renewcommand{\arraystretch}{1}
\end{table}

\begin{table}[H]
\centering
\tiny
\caption{\textbf{Half-descendant pruning stress test.} A random half of the base model descendants is pruned, the base and remaining descendants remain clean. Entries are parent-child direction accuracy after aggregating exactly $\Nw$ same-family witnesses, with $\Nw{=}\mathrm{all}$ using all eligible witnesses.}
\label{tab:pruning_nw_half_descendants}
\setlength{\tabcolsep}{2pt}
\renewcommand{\arraystretch}{0.92}
\resizebox{0.82\textwidth}{!}{%
\begin{tabular}{lllrrrr}
\toprule
Model & Rule & Magnitude pruning & $\Nw{=}1$ & $\Nw{=}2$ & $\Nw{=}3$ & $\Nw{=}\mathrm{all}$ \\
\midrule
GPT-2 & Frobenius cosine & 0\% & 87.8\% & 98.1\% & 99.8\% & 100.0\% \\
GPT-2 & Frobenius cosine & 2\% & 87.8\% & 98.1\% & 99.8\% & 100.0\% \\
GPT-2 & Frobenius cosine & 5\% & 88.9\% & 98.1\% & 99.6\% & 100.0\% \\
GPT-2 & Frobenius cosine & 10\% & 83.3\% & 92.5\% & 96.0\% & 100.0\% \\
GPT-2 & Frobenius cosine & 30\% & 62.2\% & 77.5\% & 84.9\% & 100.0\% \\
GPT-2 & Frobenius cosine & 50\% & 62.2\% & 76.1\% & 76.3\% & 80.0\% \\
\addlinespace[1pt]
GPT-2 & SVD top-$k{=}16$ & 0\% & 91.1\% & 97.2\% & 99.2\% & 100.0\% \\
GPT-2 & SVD top-$k{=}16$ & 2\% & 90.0\% & 97.5\% & 99.2\% & 100.0\% \\
GPT-2 & SVD top-$k{=}16$ & 5\% & 90.0\% & 97.2\% & 99.2\% & 100.0\% \\
GPT-2 & SVD top-$k{=}16$ & 10\% & 88.9\% & 96.9\% & 99.2\% & 100.0\% \\
GPT-2 & SVD top-$k{=}16$ & 30\% & 77.8\% & 90.6\% & 93.6\% & 100.0\% \\
GPT-2 & SVD top-$k{=}16$ & 50\% & 64.4\% & 77.5\% & 78.7\% & 80.0\% \\
\bottomrule
\end{tabular}
}
\renewcommand{\arraystretch}{1}
\end{table}

\begin{table}[H]
\centering
\tiny
\caption{\textbf{All-descendant pruning stress test.} All GPT-2 descendants are pruned, while the base checkpoint remains clean. Entries are parent-child direction accuracy after aggregating exactly $\Nw$ same-family witnesses, with $\Nw{=}\mathrm{all}$ using all eligible witnesses.}
\label{tab:pruning_nw_all_descendants}
\setlength{\tabcolsep}{2pt}
\renewcommand{\arraystretch}{0.92}
\resizebox{0.82\textwidth}{!}{%
\begin{tabular}{lllrrrr}
\toprule
Model & Rule & Magnitude pruning & $\Nw{=}1$ & $\Nw{=}2$ & $\Nw{=}3$ & $\Nw{=}\mathrm{all}$ \\
\midrule
GPT-2 & Frobenius cosine & 0\% & 87.8\% & 98.1\% & 99.8\% & 100.0\% \\
GPT-2 & Frobenius cosine & 2\% & 83.3\% & 90.3\% & 92.1\% & 100.0\% \\
GPT-2 & Frobenius cosine & 5\% & 73.3\% & 76.1\% & 76.3\% & 70.0\% \\
GPT-2 & Frobenius cosine & 10\% & 53.3\% & 52.5\% & 49.5\% & 50.0\% \\
GPT-2 & Frobenius cosine & 30\% & 17.8\% & 17.8\% & 20.0\% & 20.0\% \\
GPT-2 & Frobenius cosine & 50\% & 17.8\% & 15.6\% & 20.0\% & 20.0\% \\
\addlinespace[1pt]
GPT-2 & SVD top-$k{=}16$ & 0\% & 92.2\% & 97.5\% & 99.2\% & 100.0\% \\
GPT-2 & SVD top-$k{=}16$ & 2\% & 90.0\% & 92.8\% & 95.1\% & 100.0\% \\
GPT-2 & SVD top-$k{=}16$ & 5\% & 83.3\% & 83.3\% & 83.7\% & 80.0\% \\
GPT-2 & SVD top-$k{=}16$ & 10\% & 77.8\% & 76.1\% & 74.3\% & 70.0\% \\
GPT-2 & SVD top-$k{=}16$ & 30\% & 41.1\% & 49.2\% & 50.0\% & 50.0\% \\
GPT-2 & SVD top-$k{=}16$ & 50\% & 17.8\% & 17.8\% & 20.0\% & 20.0\% \\
\bottomrule
\end{tabular}
}
\renewcommand{\arraystretch}{1}
\end{table}

\FloatBarrier

\section{Child Gaussian Noise Stress Test}
\label{app:child_noise_stress}

We run a child-only Gaussian noise stress test as a robustness sanity check for the Frobenius-cosine direction rule. For each parent-child pair $P\to C$ and each evaluated 2D projection tensor $\ell$, define the clean child update
\[
u_\ell=\theta_C^{(\ell)}-\theta_P^{(\ell)}.
\]
We sample isotropic Gaussian noise and rescale it to
\[
\|\epsilon_\ell\|_F=\alpha\|u_\ell\|_F.
\]
Only the child checkpoint is perturbed,
\[
\widetilde{\theta}_C^{(\ell)}=\theta_C^{(\ell)}+\epsilon_\ell,
\]
while the parent and witness remain clean. We then recompute the Frobenius-cosine direction decision by comparing $C(P;\widetilde C,W)$ with $C(\widetilde C;P,W)$. The prediction is correct when the parent-anchor score is lower. For $\alpha=0$, we report the clean evaluation. For each nonzero $\alpha$, we merge results from three random seeds.

\begin{table}[H]
\centering
\small
\caption{\textbf{Child Gaussian noise stress test.} Only the child checkpoint is perturbed. Larger $\alpha$ does not reduce Frobenius-cosine direction accuracy on these two families.}
\label{tab:child_noise_stress}
\setlength{\tabcolsep}{8pt}
\begin{tabular}{rcc}
\toprule
$\alpha$ & GPT-2 & TinyLlama-1.1B-Chat-v1.0 \\
\midrule
0 & $79/90=87.8\%$ & $54/56=96.4\%$ \\
0.1 & $237/270=87.8\%$ & $162/168=96.4\%$ \\
0.2 & $237/270=87.8\%$ & $162/168=96.4\%$ \\
0.5 & $240/270=88.9\%$ & $168/168=100.0\%$ \\
1 & $246/270=91.1\%$ & $168/168=100.0\%$ \\
2 & $252/270=93.3\%$ & $168/168=100.0\%$ \\
5 & $270/270=100.0\%$ & $168/168=100.0\%$ \\
10 & $270/270=100.0\%$ & $168/168=100.0\%$ \\
\bottomrule
\end{tabular}
\end{table}

These child-side perturbations do not reduce accuracy in this setting (\Cref{tab:child_noise_stress}). Small perturbations leave the result essentially unchanged. Larger perturbations tend to make the direction easier, because the noisy child as anchor induces two deltas that share the same strong noise component, increasing child-anchor overlap. In contrast, the parent-anchor score compares the noisy child update with a clean witness update, where the added noise is nearly orthogonal to the witness direction.

\section{Per-Family Composition and Model Registry}
\label{app:rlhf_per_family}

\paragraph{Use disclaimer.}
The listed checkpoints are used solely for local academic evaluation. We do not redistribute, mirror, host, sell, sublicense, or otherwise provide model weights or checkpoint files; all access remains through the original public repositories.

\begin{center}
\tiny
\setlength{\tabcolsep}{2pt}
\begin{longtable}{>{\raggedright\arraybackslash}p{0.24\linewidth} >{\raggedright\arraybackslash}p{0.18\linewidth} >{\raggedright\arraybackslash}p{0.54\linewidth}}
\caption{\textbf{Full LLM checkpoint registry.} The table lists the public HuggingFace repositories used for local academic evaluation.}\label{tab:llm_registry}\\
\toprule
Family & Checkpoint type & HuggingFace repository \\
\midrule
\endfirsthead
\toprule
Family & Checkpoint type & HuggingFace repository \\
\midrule
\endhead
\midrule
\multicolumn{3}{r}{continued on next page}\\
\endfoot
\bottomrule
\endlastfoot
DeepSeek-R1-Distill-Qwen-1.5B & base & \texttt{deepseek-\allowbreak{}ai/\allowbreak{}DeepSeek-\allowbreak{}R1-\allowbreak{}Distill-\allowbreak{}Qwen-\allowbreak{}1.\allowbreak{}5B} \\
DeepSeek-R1-Distill-Qwen-1.5B & SFT & \texttt{jan-\allowbreak{}hq/\allowbreak{}AlphaMaze-\allowbreak{}v0.\allowbreak{}2-\allowbreak{}1.\allowbreak{}5B-\allowbreak{}SFT} \\
DeepSeek-R1-Distill-Qwen-1.5B & SFT & \texttt{Menlo/\allowbreak{}AlphaSpace-\allowbreak{}1.\allowbreak{}5B} \\
DeepSeek-R1-Distill-Qwen-1.5B & SFT & \texttt{thuml/\allowbreak{}bytesized32-\allowbreak{}world-\allowbreak{}model-\allowbreak{}sft} \\
DeepSeek-R1-Distill-Qwen-1.5B & SFT & \texttt{lightblue/\allowbreak{}DeepSeek-\allowbreak{}R1-\allowbreak{}Distill-\allowbreak{}Qwen-\allowbreak{}1.\allowbreak{}5B-\allowbreak{}Multilingual} \\
DeepSeek-R1-Distill-Qwen-1.5B & SFT & \texttt{avanishd/\allowbreak{}DeepSeek-\allowbreak{}R1-\allowbreak{}Distill-\allowbreak{}Qwen-\allowbreak{}1.\allowbreak{}5B-\allowbreak{}finetuned-\allowbreak{}smoltalk-\allowbreak{}everyday-\allowbreak{}conversations} \\
DeepSeek-R1-Distill-Qwen-1.5B & SFT & \texttt{UCSC-\allowbreak{}VLAA/\allowbreak{}STAR1-\allowbreak{}R1-\allowbreak{}Distill-\allowbreak{}1.\allowbreak{}5B} \\
DeepSeek-R1-Distill-Qwen-1.5B & SFT & \texttt{Ethencam/\allowbreak{}lora-\allowbreak{}deepseek-\allowbreak{}qwen-\allowbreak{}1.\allowbreak{}5B} \\
DeepSeek-R1-Distill-Qwen-1.5B & RLHF & \texttt{agentica-\allowbreak{}org/\allowbreak{}DeepScaleR-\allowbreak{}1.\allowbreak{}5B-\allowbreak{}Preview} \\
DeepSeek-R1-Distill-Qwen-1.5B & RLHF & \texttt{Zyphra/\allowbreak{}ZR1-\allowbreak{}1.\allowbreak{}5B} \\
Gemma-3-4B-it & base & \texttt{google/\allowbreak{}gemma-\allowbreak{}3-\allowbreak{}4b-\allowbreak{}it} \\
Gemma-3-4B-it & SFT & \texttt{INSAIT-\allowbreak{}Institute/\allowbreak{}BgGPT-\allowbreak{}Gemma-\allowbreak{}3-\allowbreak{}4B-\allowbreak{}IT} \\
Gemma-3-4B-it & SFT & \texttt{allura-\allowbreak{}org/\allowbreak{}Gemma-\allowbreak{}3-\allowbreak{}Glitter-\allowbreak{}4B} \\
Gemma-3-4B-it & SFT & \texttt{INSAIT-\allowbreak{}Institute/\allowbreak{}MamayLM-\allowbreak{}Gemma-\allowbreak{}3-\allowbreak{}4B-\allowbreak{}IT-\allowbreak{}v1.\allowbreak{}0} \\
Gemma-3-4B-it & SFT & \texttt{nvidia/\allowbreak{}Nemotron-\allowbreak{}Content-\allowbreak{}Safety-\allowbreak{}Reasoning-\allowbreak{}4B} \\
Gemma-3-4B-it & SFT & \texttt{sarvamai/\allowbreak{}sarvam-\allowbreak{}translate} \\
Gemma-3-4B-it & SFT & \texttt{neo4j/\allowbreak{}text-\allowbreak{}to-\allowbreak{}cypher-\allowbreak{}Gemma-\allowbreak{}3-\allowbreak{}4B-\allowbreak{}Instruct-\allowbreak{}2025.\allowbreak{}04.\allowbreak{}0} \\
Gemma-3-4B-it & RLHF & \texttt{anthonym21/\allowbreak{}gemma-\allowbreak{}3-\allowbreak{}4b-\allowbreak{}it-\allowbreak{}slipstream-\allowbreak{}grpo} \\
GPT-2 & base & \texttt{openai-\allowbreak{}community/\allowbreak{}gpt2} \\
GPT-2 & SFT & \texttt{nicholasKluge/\allowbreak{}Aira-\allowbreak{}2-\allowbreak{}124M} \\
GPT-2 & SFT & \texttt{juancopi81/\allowbreak{}gpt2-\allowbreak{}finetuned-\allowbreak{}yannic-\allowbreak{}test} \\
GPT-2 & SFT & \texttt{harigovind511/\allowbreak{}GPT2-\allowbreak{}Guanaco-\allowbreak{}LoRA} \\
GPT-2 & SFT & \texttt{Ssarion/\allowbreak{}gpt2-\allowbreak{}multi-\allowbreak{}news} \\
GPT-2 & SFT & \texttt{kpriyanshu256/\allowbreak{}gpt-\allowbreak{}ya2-\allowbreak{}v2} \\
GPT-2 & SFT & \texttt{bipin/\allowbreak{}malayalam-\allowbreak{}gpt2} \\
GPT-2 & SFT & \texttt{alibidaran/\allowbreak{}medical\_\allowbreak{}transcription\_\allowbreak{}generator} \\
GPT-2 & SFT & \texttt{jackoyoungblood/\allowbreak{}TinyStoriesTest} \\
GPT-2 & SFT & \texttt{crumb/\allowbreak{}qrstudy-\allowbreak{}gpt2-\allowbreak{}2-\allowbreak{}4} \\
GPT-2 & SFT & \texttt{Winmodel/\allowbreak{}gpt2-\allowbreak{}lora-\allowbreak{}aligned-\allowbreak{}orpo} \\
GPT-2 Medium & base & \texttt{openai-\allowbreak{}community/\allowbreak{}gpt2-\allowbreak{}medium} \\
GPT-2 Medium & SFT & \texttt{nicholasKluge/\allowbreak{}Aira-\allowbreak{}2-\allowbreak{}355M} \\
GPT-2 Medium & SFT & \texttt{samaksh-\allowbreak{}khatri-\allowbreak{}crest-\allowbreak{}data/\allowbreak{}gmra\_\allowbreak{}model\_\allowbreak{}gpt2-\allowbreak{}medium\_\allowbreak{}14082023T134929} \\
GPT-2 Medium & SFT & \texttt{shubhamgantayat/\allowbreak{}gpt2-\allowbreak{}medium-\allowbreak{}custom} \\
GPT-2 Medium & SFT & \texttt{ckandemir/\allowbreak{}gpt2-\allowbreak{}medium-\allowbreak{}finetuned-\allowbreak{}contract-\allowbreak{}gen} \\
GPT-2 Medium & SFT & \texttt{umaru97/\allowbreak{}gpt2-\allowbreak{}product-\allowbreak{}review-\allowbreak{}generation} \\
GPT-2 Medium & SFT & \texttt{jkhan447/\allowbreak{}results} \\
GPT-2 Medium & SFT & \texttt{kunjcr2/\allowbreak{}gpt-\allowbreak{}lora} \\
GPT-2 Medium & SFT & \texttt{Salm00n/\allowbreak{}gpt2-\allowbreak{}medium\_\allowbreak{}RACE-\allowbreak{}H\_\allowbreak{}v1} \\
GPT-2 Medium & RLHF & \texttt{gumran/\allowbreak{}gpt2-\allowbreak{}medium-\allowbreak{}dpo} \\
GPT-2 Medium & RLHF & \texttt{itsmepv/\allowbreak{}gpt2-\allowbreak{}ppo-\allowbreak{}prefix-\allowbreak{}tuning} \\
GPT-2 XL & base & \texttt{openai-\allowbreak{}community/\allowbreak{}gpt2-\allowbreak{}xl} \\
GPT-2 XL & SFT & \texttt{masani/\allowbreak{}2025-\allowbreak{}04-\allowbreak{}02\_\allowbreak{}14-\allowbreak{}52-\allowbreak{}39} \\
GPT-2 XL & SFT & \texttt{santis2/\allowbreak{}gpt2-\allowbreak{}xl-\allowbreak{}alpaca-\allowbreak{}instruction-\allowbreak{}fine-\allowbreak{}tuning-\allowbreak{}lora} \\
GPT-2 XL & SFT & \texttt{aieng-\allowbreak{}lab/\allowbreak{}gpt2-\allowbreak{}xl\_\allowbreak{}sentiment} \\
GPT-2 XL & SFT & \texttt{styalai/\allowbreak{}gpt2o-\allowbreak{}chatbot-\allowbreak{}02} \\
GPT-2 XL & SFT & \texttt{masani/\allowbreak{}sft\_\allowbreak{}checkpoints} \\
GPT-2 XL & SFT & \texttt{MiniLLM/\allowbreak{}teacher-\allowbreak{}gpt2-\allowbreak{}1.\allowbreak{}5B} \\
GPT-2 XL & SFT & \texttt{MHGanainy/\allowbreak{}gpt2-\allowbreak{}xl-\allowbreak{}lora-\allowbreak{}multi-\allowbreak{}512-\allowbreak{}3-\allowbreak{}top} \\
GPT-2 XL & SFT & \texttt{manueldeprada/\allowbreak{}gflownet-\allowbreak{}next-\allowbreak{}sentence-\allowbreak{}gpt2xl-\allowbreak{}0.\allowbreak{}90} \\
Llama-3.2-1B-Instruct & base & \texttt{meta-\allowbreak{}llama/\allowbreak{}Llama-\allowbreak{}3.\allowbreak{}2-\allowbreak{}1B-\allowbreak{}Instruct} \\
Llama-3.2-1B-Instruct & SFT & \texttt{AvaLovelace/\allowbreak{}BrickGPT} \\
Llama-3.2-1B-Instruct & SFT & \texttt{minpeter/\allowbreak{}LoRA-\allowbreak{}Llama-\allowbreak{}3.\allowbreak{}2-\allowbreak{}1B-\allowbreak{}tool-\allowbreak{}vllm-\allowbreak{}ci} \\
Llama-3.2-1B-Instruct & SFT & \texttt{erbacher/\allowbreak{}llama3.\allowbreak{}2-\allowbreak{}1B-\allowbreak{}MATH} \\
Llama-3.2-1B-Instruct & SFT & \texttt{alpha-\allowbreak{}ai/\allowbreak{}Medical-\allowbreak{}Guide-\allowbreak{}COT-\allowbreak{}llama3.\allowbreak{}2-\allowbreak{}1B} \\
Llama-3.2-1B-Instruct & SFT & \texttt{Aygun/\allowbreak{}llama-\allowbreak{}3.\allowbreak{}2-\allowbreak{}1B-\allowbreak{}MLQRECC-\allowbreak{}Rewriter} \\
Llama-3.2-1B-Instruct & SFT & \texttt{diabolic6045/\allowbreak{}open-\allowbreak{}llama-\allowbreak{}3.\allowbreak{}2-\allowbreak{}1B-\allowbreak{}Instruct} \\
Llama-3.2-1B-Instruct & SFT & \texttt{azizmatin/\allowbreak{}llama3.\allowbreak{}2-\allowbreak{}1B-\allowbreak{}persianQAV2.\allowbreak{}0} \\
Llama-3.2-1B-Instruct & SFT & \texttt{syvai/\allowbreak{}reasoning-\allowbreak{}gen-\allowbreak{}1b} \\
Llama-3.2-1B-Instruct & SFT & \texttt{minmax23/\allowbreak{}llama3-\allowbreak{}2-\allowbreak{}1b-\allowbreak{}lora} \\
Llama-3.2-1B-Instruct & SFT & \texttt{burtenshaw/\allowbreak{}Llama-\allowbreak{}3.\allowbreak{}2-\allowbreak{}1B-\allowbreak{}Tulu3-\allowbreak{}LoRA} \\
Llama-3.2-1B-Instruct & RLHF & \texttt{ahczhg/\allowbreak{}Llama-\allowbreak{}3.\allowbreak{}2-\allowbreak{}1B-\allowbreak{}Aegis-\allowbreak{}SFT-\allowbreak{}DPO} \\
Llama-3.2-1B-Instruct & RLHF & \texttt{BleachNick/\allowbreak{}Llama-\allowbreak{}3.\allowbreak{}2-\allowbreak{}1B-\allowbreak{}Instruct-\allowbreak{}GRPO-\allowbreak{}45k\_\allowbreak{}RAGv2} \\
Llama-3.2-1B-Instruct & RLHF & \texttt{pranav6905/\allowbreak{}llama-\allowbreak{}1b-\allowbreak{}sft-\allowbreak{}dpo-\allowbreak{}final} \\
Llama-3.2-1B-Instruct & RLHF & \texttt{ericflo/\allowbreak{}Llama-\allowbreak{}3.\allowbreak{}2-\allowbreak{}1B-\allowbreak{}Instruct-\allowbreak{}RLHF-\allowbreak{}v0.\allowbreak{}1} \\
Meta-Llama-3-8B & base & \texttt{meta-\allowbreak{}llama/\allowbreak{}Meta-\allowbreak{}Llama-\allowbreak{}3-\allowbreak{}8B} \\
Meta-Llama-3-8B & SFT & \texttt{Weyaxi/\allowbreak{}Einstein-\allowbreak{}v6.\allowbreak{}1-\allowbreak{}Llama3-\allowbreak{}8B} \\
Meta-Llama-3-8B & SFT & \texttt{DeepMount00/\allowbreak{}Llama-\allowbreak{}3-\allowbreak{}8b-\allowbreak{}Ita} \\
Meta-Llama-3-8B & SFT & \texttt{Magpie-\allowbreak{}Align/\allowbreak{}Llama-\allowbreak{}3-\allowbreak{}8B-\allowbreak{}Magpie-\allowbreak{}Align-\allowbreak{}SFT-\allowbreak{}v0.\allowbreak{}3} \\
Meta-Llama-3-8B & SFT & \texttt{openchat/\allowbreak{}openchat-\allowbreak{}3.\allowbreak{}6-\allowbreak{}8b-\allowbreak{}20240522} \\
Meta-Llama-3-8B & SFT & \texttt{OpenRLHF/\allowbreak{}Llama-\allowbreak{}3-\allowbreak{}8b-\allowbreak{}sft-\allowbreak{}mixture} \\
Meta-Llama-3-8B & SFT & \texttt{YipingZhang/\allowbreak{}Meta-\allowbreak{}Llama-\allowbreak{}3-\allowbreak{}8B} \\
Meta-Llama-3-8B & SFT & \texttt{Siqi-\allowbreak{}Hu/\allowbreak{}Llama3-\allowbreak{}8B-\allowbreak{}lora-\allowbreak{}r-\allowbreak{}32-\allowbreak{}generic-\allowbreak{}step-\allowbreak{}1200-\allowbreak{}lr-\allowbreak{}1e-\allowbreak{}5-\allowbreak{}labels\_\allowbreak{}40.\allowbreak{}0-\allowbreak{}1} \\
Meta-Llama-3-8B & SFT & \texttt{EleutherAI/\allowbreak{}Meta-\allowbreak{}Llama-\allowbreak{}3-\allowbreak{}8B-\allowbreak{}sciq-\allowbreak{}random-\allowbreak{}standardized-\allowbreak{}random-\allowbreak{}names} \\
Meta-Llama-3-8B & RLHF & \texttt{dfurman/\allowbreak{}Llama-\allowbreak{}3-\allowbreak{}8B-\allowbreak{}Orpo-\allowbreak{}v0.\allowbreak{}1} \\
Meta-Llama-3-8B & RLHF & \texttt{mlabonne/\allowbreak{}OrpoLlama-\allowbreak{}3-\allowbreak{}8B} \\
Mistral-7B-v0.3 & base & \texttt{mistralai/\allowbreak{}Mistral-\allowbreak{}7B-\allowbreak{}v0.\allowbreak{}3} \\
Mistral-7B-v0.3 & SFT & \texttt{pucpr-\allowbreak{}br/\allowbreak{}Clinical-\allowbreak{}BR-\allowbreak{}Mistral-\allowbreak{}7B-\allowbreak{}v0.\allowbreak{}2} \\
Mistral-7B-v0.3 & SFT & \texttt{cypienai/\allowbreak{}cymist-\allowbreak{}2-\allowbreak{}v03-\allowbreak{}SFT} \\
Mistral-7B-v0.3 & SFT & \texttt{dphn/\allowbreak{}dolphin-\allowbreak{}2.\allowbreak{}9.\allowbreak{}3-\allowbreak{}mistral-\allowbreak{}7B-\allowbreak{}32k} \\
Mistral-7B-v0.3 & SFT & \texttt{OpenLLM-\allowbreak{}Ro/\allowbreak{}RoMistral-\allowbreak{}7b-\allowbreak{}Instruct} \\
Mistral-7B-v0.3 & SFT & \texttt{openfoodfacts/\allowbreak{}spellcheck-\allowbreak{}mistral-\allowbreak{}7b} \\
Mistral-7B-v0.3 & SFT & \texttt{migtissera/\allowbreak{}Tess-\allowbreak{}3-\allowbreak{}7B-\allowbreak{}SFT} \\
Mistral-7B-v0.3 & SFT & \texttt{yspkm/\allowbreak{}Mistral-\allowbreak{}7B-\allowbreak{}v0.\allowbreak{}3-\allowbreak{}lora-\allowbreak{}math} \\
Mistral-7B-v0.3 & SFT & \texttt{TatarNLPWorld/\allowbreak{}mistral-\allowbreak{}7b-\allowbreak{}tatar-\allowbreak{}lora-\allowbreak{}r8} \\
Mistral-7B-v0.3 & RLHF & \texttt{shenzhi-\allowbreak{}wang/\allowbreak{}Mistral-\allowbreak{}7B-\allowbreak{}v0.\allowbreak{}3-\allowbreak{}Chinese-\allowbreak{}Chat} \\
Mistral-7B-v0.3 & RLHF & \texttt{Ppear/\allowbreak{}mistral7b-\allowbreak{}grpo} \\
Pythia-410M & base & \texttt{EleutherAI/\allowbreak{}pythia-\allowbreak{}410m} \\
Pythia-410M & SFT & \texttt{vandijklab/\allowbreak{}C2S-\allowbreak{}Pythia-\allowbreak{}410m-\allowbreak{}cell-\allowbreak{}type-\allowbreak{}prediction} \\
Pythia-410M & SFT & \texttt{nnheui/\allowbreak{}pythia-\allowbreak{}410m-\allowbreak{}sft-\allowbreak{}full} \\
Pythia-410M & SFT & \texttt{Joshua-\allowbreak{}Abok/\allowbreak{}Malawi-\allowbreak{}Public-\allowbreak{}Health-\allowbreak{}Systems} \\
Pythia-410M & SFT & \texttt{kykim0/\allowbreak{}pythia-\allowbreak{}410m-\allowbreak{}tulu-\allowbreak{}v2-\allowbreak{}mix} \\
Pythia-410M & SFT & \texttt{mia-\allowbreak{}llm/\allowbreak{}pythia-\allowbreak{}410m-\allowbreak{}xsum-\allowbreak{}oneepoch} \\
Pythia-410M & RLHF & \texttt{lomahony/\allowbreak{}pythia-\allowbreak{}410m-\allowbreak{}helpful-\allowbreak{}dpo} \\
Pythia-410M & RLHF & \texttt{lomahony/\allowbreak{}eleuther-\allowbreak{}pythia410m-\allowbreak{}hh-\allowbreak{}dpo} \\
Qwen2.5-0.5B & base & \texttt{Qwen/\allowbreak{}Qwen2.\allowbreak{}5-\allowbreak{}0.\allowbreak{}5B} \\
Qwen2.5-0.5B & SFT & \texttt{robertvacareanu/\allowbreak{}instruct-\allowbreak{}rl-\allowbreak{}Qwen2.\allowbreak{}5-\allowbreak{}0.\allowbreak{}5B-\allowbreak{}databricks-\allowbreak{}dolly-\allowbreak{}15k} \\
Qwen2.5-0.5B & SFT & \texttt{evalstate/\allowbreak{}qwen-\allowbreak{}demo-\allowbreak{}verbose} \\
Qwen2.5-0.5B & SFT & \texttt{JayHyeon/\allowbreak{}Qwen2.\allowbreak{}5-\allowbreak{}0.\allowbreak{}5B-\allowbreak{}SFT} \\
Qwen2.5-0.5B & SFT & \texttt{qgallouedec/\allowbreak{}Qwen2.\allowbreak{}5-\allowbreak{}0.\allowbreak{}5B-\allowbreak{}SFT} \\
Qwen2.5-0.5B & SFT & \texttt{lewtun/\allowbreak{}Qwen2.\allowbreak{}5-\allowbreak{}0.\allowbreak{}5B-\allowbreak{}SFT-\allowbreak{}LoRA} \\
Qwen2.5-0.5B & SFT & \texttt{artificialguybr/\allowbreak{}QWEN-\allowbreak{}2.\allowbreak{}5-\allowbreak{}0.\allowbreak{}5B-\allowbreak{}Synthia-\allowbreak{}I} \\
Qwen2.5-0.5B & SFT & \texttt{evalstate/\allowbreak{}trl-\allowbreak{}demo-\allowbreak{}qwen} \\
Qwen2.5-0.5B & SFT & \texttt{rahul7star/\allowbreak{}Qwen2.\allowbreak{}5-\allowbreak{}0.\allowbreak{}5B-\allowbreak{}Gita} \\
Qwen2.5-0.5B & SFT & \texttt{VERSIL91/\allowbreak{}6d762c66-\allowbreak{}b841-\allowbreak{}4dba-\allowbreak{}92e3-\allowbreak{}ceddc0b67beb} \\
Qwen2.5-0.5B & RLHF & \texttt{shawhin/\allowbreak{}Qwen2.\allowbreak{}5-\allowbreak{}0.\allowbreak{}5B-\allowbreak{}DPO} \\
Qwen2.5-0.5B & RLHF & \texttt{Saralatifi/\allowbreak{}Qwen2.\allowbreak{}5-\allowbreak{}0.\allowbreak{}5B-\allowbreak{}DPO} \\
Qwen2.5-0.5B & RLHF & \texttt{BounharAbdelaziz/\allowbreak{}Qwen2.\allowbreak{}5-\allowbreak{}0.\allowbreak{}5B-\allowbreak{}DPO-\allowbreak{}English-\allowbreak{}Orca} \\
Qwen2.5-0.5B & RLHF & \texttt{jaygala24/\allowbreak{}Qwen2.\allowbreak{}5-\allowbreak{}0.\allowbreak{}5B-\allowbreak{}GRPO-\allowbreak{}math-\allowbreak{}reasoning} \\
Qwen2.5-0.5B & RLHF & \texttt{XueyingJia/\allowbreak{}qwen2.\allowbreak{}5-\allowbreak{}0.\allowbreak{}5b-\allowbreak{}oaif} \\
Qwen3-0.6B-Base & base & \texttt{Qwen/\allowbreak{}Qwen3-\allowbreak{}0.\allowbreak{}6B-\allowbreak{}Base} \\
Qwen3-0.6B-Base & SFT & \texttt{mrfakename/\allowbreak{}dreamwriter-\allowbreak{}0.\allowbreak{}6b-\allowbreak{}beta} \\
Qwen3-0.6B-Base & SFT & \texttt{EhDa24/\allowbreak{}MNLP\_\allowbreak{}M2\_\allowbreak{}mcqa\_\allowbreak{}model\_\allowbreak{}full\_\allowbreak{}ft} \\
Qwen3-0.6B-Base & SFT & \texttt{cam-\allowbreak{}1000/\allowbreak{}MNLP\_\allowbreak{}M2\_\allowbreak{}rag\_\allowbreak{}model} \\
Qwen3-0.6B-Base & SFT & \texttt{intalio/\allowbreak{}Qwen3-\allowbreak{}NoThink-\allowbreak{}0.\allowbreak{}6b} \\
Qwen3-0.6B-Base & SFT & \texttt{matyaydin/\allowbreak{}rag\_\allowbreak{}old} \\
Qwen3-0.6B-Base & SFT & \texttt{charlottemeyer/\allowbreak{}s1-\allowbreak{}20250515\_\allowbreak{}101249} \\
Qwen3-0.6B-Base & SFT & \texttt{LastTransformer/\allowbreak{}qlora-\allowbreak{}qwen3-\allowbreak{}0.\allowbreak{}6B-\allowbreak{}lora-\allowbreak{}dpo} \\
Qwen3-0.6B-Base & SFT & \texttt{ldqvinh/\allowbreak{}qwen3-\allowbreak{}0.\allowbreak{}6b-\allowbreak{}base-\allowbreak{}grpo-\allowbreak{}v2-\allowbreak{}2048} \\
Qwen3-0.6B-Base & RLHF & \texttt{AIPlans/\allowbreak{}Qwen3-\allowbreak{}0.\allowbreak{}6B-\allowbreak{}IPO} \\
Qwen3-8B & base & \texttt{Qwen/\allowbreak{}Qwen3-\allowbreak{}8B} \\
Qwen3-8B & SFT & \texttt{rl-\allowbreak{}research/\allowbreak{}DR-\allowbreak{}Tulu-\allowbreak{}SFT-\allowbreak{}8B} \\
Qwen3-8B & SFT & \texttt{Goedel-\allowbreak{}LM/\allowbreak{}Goedel-\allowbreak{}Prover-\allowbreak{}V2-\allowbreak{}8B} \\
Qwen3-8B & SFT & \texttt{open-\allowbreak{}thoughts/\allowbreak{}OpenThinker-\allowbreak{}Agent-\allowbreak{}v1-\allowbreak{}SFT} \\
Qwen3-8B & SFT & \texttt{nvidia/\allowbreak{}Nemotron-\allowbreak{}Orchestrator-\allowbreak{}8B} \\
Qwen3-8B & SFT & \texttt{shisa-\allowbreak{}ai/\allowbreak{}shisa-\allowbreak{}v2.\allowbreak{}1-\allowbreak{}qwen3-\allowbreak{}8b} \\
Qwen3-8B & SFT & \texttt{xinli95/\allowbreak{}Qwen3-\allowbreak{}8B-\allowbreak{}SFT} \\
Qwen3-8B & RLHF & \texttt{open-\allowbreak{}thoughts/\allowbreak{}OpenThinker-\allowbreak{}Agent-\allowbreak{}v1} \\
SmolLM2-135M & base & \texttt{HuggingFaceTB/\allowbreak{}SmolLM2-\allowbreak{}135M} \\
SmolLM2-135M & SFT & \texttt{Ellight/\allowbreak{}code-\allowbreak{}smolLM2-\allowbreak{}135m-\allowbreak{}text-\allowbreak{}to-\allowbreak{}sql} \\
SmolLM2-135M & SFT & \texttt{gotoplanb/\allowbreak{}SmolLM2-\allowbreak{}FT-\allowbreak{}MyDataset} \\
SmolLM2-135M & SFT & \texttt{Nels2/\allowbreak{}SmolLM2-\allowbreak{}FT-\allowbreak{}SQL-\allowbreak{}Context} \\
SmolLM2-135M & SFT & \texttt{puettmann/\allowbreak{}SmolLM2-\allowbreak{}135M-\allowbreak{}Instruct-\allowbreak{}Smol-\allowbreak{}Course} \\
SmolLM2-135M & SFT & \texttt{HuggingFaceTB/\allowbreak{}smollm2-\allowbreak{}135M-\allowbreak{}SFT-\allowbreak{}Only} \\
SmolLM2-135M & SFT & \texttt{ParitKansal/\allowbreak{}SmolLM2-\allowbreak{}135M-\allowbreak{}SFT-\allowbreak{}smoltalk} \\
SmolLM2-135M & SFT & \texttt{Ashed00/\allowbreak{}SmolMath-\allowbreak{}135M} \\
SmolLM2-135M & SFT & \texttt{Ellight/\allowbreak{}SmolLM2-\allowbreak{}135M-\allowbreak{}text-\allowbreak{}to-\allowbreak{}sql} \\
SmolLM2-135M & SFT & \texttt{mnoukhov/\allowbreak{}SmolLM2-\allowbreak{}135M-\allowbreak{}tldr-\allowbreak{}sft} \\
SmolLM2-135M & SFT & \texttt{lhoestq/\allowbreak{}finetune\_\allowbreak{}smollm2\_\allowbreak{}python} \\
SmolLM2-135M & SFT & \texttt{frankenstein-\allowbreak{}ai/\allowbreak{}admin-\allowbreak{}SmolLM2-\allowbreak{}135M-\allowbreak{}20251204\_\allowbreak{}194139} \\
SmolLM2-135M & RLHF & \texttt{MUTSC/\allowbreak{}SmolLM2-\allowbreak{}FT-\allowbreak{}ORPO} \\
SmolLM2-135M & RLHF & \texttt{hsila/\allowbreak{}SmolLM2-\allowbreak{}135M-\allowbreak{}ORPO} \\
SmolLM2-135M & RLHF & \texttt{peaceAsh/\allowbreak{}smolcourse\_\allowbreak{}chapter2\_\allowbreak{}ORPO} \\
SmolLM2-1.7B-Instruct & base & \texttt{HuggingFaceTB/\allowbreak{}SmolLM2-\allowbreak{}1.\allowbreak{}7B-\allowbreak{}Instruct} \\
SmolLM2-1.7B-Instruct & SFT & \texttt{m-\allowbreak{}ric/\allowbreak{}OpenR1-\allowbreak{}SmolLM2-\allowbreak{}1.\allowbreak{}7B-\allowbreak{}Instruct-\allowbreak{}Agentic} \\
SmolLM2-1.7B-Instruct & SFT & \texttt{motexture/\allowbreak{}SmolLCoder-\allowbreak{}1.\allowbreak{}7B-\allowbreak{}Instruct} \\
SmolLM2-1.7B-Instruct & SFT & \texttt{AssistantsLab/\allowbreak{}SmolLM2-\allowbreak{}1.\allowbreak{}7B-\allowbreak{}humanized} \\
SmolLM2-1.7B-Instruct & SFT & \texttt{47z/\allowbreak{}SmolLM2-\allowbreak{}1.\allowbreak{}7B-\allowbreak{}mgsm-\allowbreak{}math} \\
SmolLM2-1.7B-Instruct & SFT & \texttt{puettmann/\allowbreak{}SmolMaestra-\allowbreak{}1.\allowbreak{}7b-\allowbreak{}Translation} \\
SmolLM2-1.7B-Instruct & SFT & \texttt{thirdeyeai/\allowbreak{}SmolLM2-\allowbreak{}1.\allowbreak{}7B-\allowbreak{}Instruct-\allowbreak{}Uncensored} \\
SmolLM2-1.7B-Instruct & SFT & \texttt{eternis/\allowbreak{}SmolLM2-\allowbreak{}1.\allowbreak{}7B-\allowbreak{}lora-\allowbreak{}anonymizer} \\
SmolLM2-1.7B-Instruct & RLHF & \texttt{DSTI/\allowbreak{}DS-\allowbreak{}RLHF-\allowbreak{}1.\allowbreak{}7B} \\
SmolLM2-1.7B-Instruct & RLHF & \texttt{dleemiller/\allowbreak{}Penny-\allowbreak{}1.\allowbreak{}7B} \\
SmolLM2-360M-Instruct & base & \texttt{HuggingFaceTB/\allowbreak{}SmolLM2-\allowbreak{}360M-\allowbreak{}Instruct} \\
SmolLM2-360M-Instruct & SFT & \texttt{Michaelj1/\allowbreak{}INSTRUCT\_\allowbreak{}smolLM2-\allowbreak{}360M-\allowbreak{}finetuned-\allowbreak{}wikitext2-\allowbreak{}raw-\allowbreak{}v1} \\
SmolLM2-360M-Instruct & SFT & \texttt{mrs83/\allowbreak{}Kurtis-\allowbreak{}SmolLM2-\allowbreak{}360M-\allowbreak{}Instruct} \\
SmolLM2-360M-Instruct & SFT & \texttt{motexture/\allowbreak{}SmolLCoder-\allowbreak{}360M-\allowbreak{}Instruct} \\
SmolLM2-360M-Instruct & SFT & \texttt{sofyanwaldy/\allowbreak{}SmolLM2-\allowbreak{}360M-\allowbreak{}Instruct-\allowbreak{}finetuned} \\
SmolLM2-360M-Instruct & SFT & \texttt{belyakoff/\allowbreak{}SmolLM2-\allowbreak{}360M-\allowbreak{}Instruct-\allowbreak{}FT} \\
SmolLM2-360M-Instruct & SFT & \texttt{juniorVision/\allowbreak{}SmolLM2-\allowbreak{}360M-\allowbreak{}Instruct\_\allowbreak{}inst-\allowbreak{}sample-\allowbreak{}240809} \\
SmolLM2-360M-Instruct & SFT & \texttt{thatupiso/\allowbreak{}SmolLM2-\allowbreak{}360M-\allowbreak{}Instruct-\allowbreak{}K12-\allowbreak{}5000} \\
SmolLM2-360M-Instruct & SFT & \texttt{TheBlueObserver/\allowbreak{}SmolLM2-\allowbreak{}360M-\allowbreak{}Instruct-\allowbreak{}MLX} \\
SmolLM2-360M-Instruct & SFT & \texttt{ThatsGroes/\allowbreak{}SmolLM2-\allowbreak{}360M-\allowbreak{}Instruct-\allowbreak{}summarizer} \\
SmolLM2-360M-Instruct & SFT & \texttt{prithivMLmods/\allowbreak{}SmolLM2-\allowbreak{}CoT-\allowbreak{}360M} \\
SmolLM2-360M-Instruct & SFT & \texttt{GawdSB/\allowbreak{}story\_\allowbreak{}model} \\
SmolLM2-360M-Instruct & SFT & \texttt{Qurtana/\allowbreak{}SmolLM2-\allowbreak{}360M-\allowbreak{}Instruct-\allowbreak{}Reasoning-\allowbreak{}v1-\allowbreak{}LoRA} \\
SmolLM2-360M-Instruct & SFT & \texttt{nadaashraff/\allowbreak{}smollm2-\allowbreak{}360m-\allowbreak{}alpaca-\allowbreak{}lora} \\
SmolLM2-360M-Instruct & RLHF & \texttt{ericlewis/\allowbreak{}infinite-\allowbreak{}craft-\allowbreak{}smollm2-\allowbreak{}360m-\allowbreak{}full-\allowbreak{}grpo} \\
TinyLlama-1.1B-Chat-v1.0 & base & \texttt{TinyLlama/\allowbreak{}TinyLlama-\allowbreak{}1.\allowbreak{}1B-\allowbreak{}Chat-\allowbreak{}v1.\allowbreak{}0} \\
TinyLlama-1.1B-Chat-v1.0 & SFT & \texttt{acon96/\allowbreak{}Home-\allowbreak{}1B-\allowbreak{}v3-\allowbreak{}GGUF} \\
TinyLlama-1.1B-Chat-v1.0 & SFT & \texttt{ShieldX/\allowbreak{}manovyadh-\allowbreak{}1.\allowbreak{}1B-\allowbreak{}v1-\allowbreak{}chat} \\
TinyLlama-1.1B-Chat-v1.0 & SFT & \texttt{h4rz3rk4s3/\allowbreak{}TinyNewsLlama-\allowbreak{}1.\allowbreak{}1B} \\
TinyLlama-1.1B-Chat-v1.0 & SFT & \texttt{h4rz3rk4s3/\allowbreak{}TinyParlaMintLlama-\allowbreak{}1.\allowbreak{}1B} \\
TinyLlama-1.1B-Chat-v1.0 & SFT & \texttt{alexredna/\allowbreak{}TinyLlama-\allowbreak{}1.\allowbreak{}1B-\allowbreak{}Chat-\allowbreak{}v1.\allowbreak{}0-\allowbreak{}reasoning-\allowbreak{}v2} \\
TinyLlama-1.1B-Chat-v1.0 & SFT & \texttt{sahil239/\allowbreak{}chatbot-\allowbreak{}v2} \\
TinyLlama-1.1B-Chat-v1.0 & RLHF & \texttt{davanstrien/\allowbreak{}TinyLlama-\allowbreak{}1.\allowbreak{}1B-\allowbreak{}Chat-\allowbreak{}v1.\allowbreak{}0-\allowbreak{}intel-\allowbreak{}dpo} \\
TinyLlama-1.1B-Chat-v1.0 & RLHF & \texttt{youndukn/\allowbreak{}TinyLlama-\allowbreak{}1.\allowbreak{}1B-\allowbreak{}Chat-\allowbreak{}v1.\allowbreak{}0-\allowbreak{}reasoning-\allowbreak{}v2-\allowbreak{}dpo-\allowbreak{}romantic} \\
\end{longtable}
\end{center}

\Cref{tab:llm_registry} lists the publicly released HuggingFace repositories used for local academic evaluation. Checkpoint type indicates whether the repository is the family base, SFT, or RLHF/alignment.

\begin{table}[t]
\centering
\scriptsize
\caption{\textbf{Per-family LLM composition.} The table lists each base family, its parameter scale, and the number of SFT and RLHF descendants used in the LLM evaluation.}
\label{tab:rlhf_per_family}
\setlength{\tabcolsep}{3pt}
\begin{tabular}{l c c c}
\toprule
Family & Params & \#SFT & \#RLHF \\
\midrule
\mbox{DeepSeek-R1-Distill-Qwen-1.5B} & 1.5B & 7 & 2 \\
\mbox{Gemma-3-4B-it} & 4B & 6 & 1 \\
\mbox{GPT-2} & 124M & 10 & 0 \\
\mbox{GPT-2 Medium} & 355M & 8 & 2 \\
\mbox{GPT-2 XL} & 1.5B & 8 & 0 \\
\mbox{Llama-3.2-1B-Instruct} & 1B & 10 & 4 \\
\mbox{Meta-Llama-3-8B} & 8B & 8 & 2 \\
\mbox{Mistral-7B-v0.3} & 7B & 8 & 2 \\
\mbox{Pythia-410M} & 410M & 5 & 2 \\
\mbox{Qwen2.5-0.5B} & 0.5B & 9 & 5 \\
\mbox{Qwen3-0.6B-Base} & 0.6B & 8 & 1 \\
\mbox{Qwen3-8B} & 8B & 6 & 1 \\
\mbox{SmolLM2-135M} & 135M & 11 & 3 \\
\mbox{SmolLM2-1.7B-Instruct} & 1.7B & 7 & 2 \\
\mbox{SmolLM2-360M-Instruct} & 360M & 13 & 1 \\
\mbox{TinyLlama-1.1B-Chat-v1.0} & 1.1B & 6 & 2 \\
\midrule
\textbf{Total} & n/a & \textbf{130} & \textbf{30} \\
\bottomrule
\end{tabular}
\end{table}

\Cref{tab:rlhf_per_family} reports the per-family composition of the LLM evaluation set. Family sizes range from GPT-2 at $124$M parameters to Meta-Llama-3-8B and Qwen3-8B. Each family has one base checkpoint and between $5$ and $14$ post-training children, contributing $130$ SFT and $30$ RLHF descendants in total.

\begin{center}
\tiny
\setlength{\tabcolsep}{2pt}
\begin{longtable}{>{\raggedright\arraybackslash}p{0.11\linewidth} >{\raggedright\arraybackslash}p{0.22\linewidth} >{\raggedright\arraybackslash}p{0.16\linewidth} >{\raggedright\arraybackslash}p{0.47\linewidth}}
\caption{\textbf{Full VLM and Diffusion checkpoint registry.} The table lists the public HuggingFace repositories used for the VLM and Diffusion checks.}\label{tab:vlm_diffusion_registry}\\
\toprule
Domain & Family & Checkpoint type & HuggingFace repository \\
\midrule
\endfirsthead
\toprule
Domain & Family & Checkpoint type & HuggingFace repository \\
\midrule
\endhead
\midrule
\multicolumn{4}{r}{continued on next page}\\
\endfoot
\bottomrule
\endlastfoot
VLM & PaliGemma2-3B-pt-224 & base & \texttt{google/\allowbreak{}paligemma2-\allowbreak{}3b-\allowbreak{}pt-\allowbreak{}224} \\
VLM & PaliGemma2-3B-pt-224 & SFT & \texttt{IPEC-\allowbreak{}COMMUNITY/\allowbreak{}spatialvla-\allowbreak{}4b-\allowbreak{}224-\allowbreak{}pt} \\
VLM & PaliGemma2-3B-pt-224 & SFT & \texttt{mateoguaman/\allowbreak{}vamos} \\
VLM & PaliGemma2-3B-pt-224 & SFT & \texttt{mateoguaman/\allowbreak{}vamos\_\allowbreak{}navigation\_\allowbreak{}only} \\
VLM & PaliGemma2-3B-pt-224 & SFT & \texttt{oliveirabruno01/\allowbreak{}count\_\allowbreak{}intersection-\allowbreak{}ft-\allowbreak{}paligemma2-\allowbreak{}3b-\allowbreak{}pt-\allowbreak{}224} \\
VLM & Qwen2.5-VL-3B-Instruct & base & \texttt{Qwen/\allowbreak{}Qwen2.\allowbreak{}5-\allowbreak{}VL-\allowbreak{}3B-\allowbreak{}Instruct} \\
VLM & Qwen2.5-VL-3B-Instruct & SFT & \texttt{nanonets/\allowbreak{}Nanonets-\allowbreak{}OCR2-\allowbreak{}3B} \\
VLM & Qwen2.5-VL-3B-Instruct & SFT & \texttt{openbmb/\allowbreak{}EVisRAG-\allowbreak{}3B} \\
VLM & Qwen2.5-VL-3B-Instruct & SFT & \texttt{UCSC-\allowbreak{}VLAA/\allowbreak{}MedVLThinker-\allowbreak{}3B-\allowbreak{}RL\_\allowbreak{}m23k} \\
VLM & Qwen2.5-VL-3B-Instruct & SFT & \texttt{InternRobotics/\allowbreak{}InternVLA-\allowbreak{}M1} \\
VLM & Qwen2-VL-2B & base & \texttt{Qwen/\allowbreak{}Qwen2-\allowbreak{}VL-\allowbreak{}2B} \\
VLM & Qwen2-VL-2B & SFT & \texttt{turningpoint-\allowbreak{}ai/\allowbreak{}VisualThinker-\allowbreak{}R1-\allowbreak{}Zero} \\
VLM & Qwen2-VL-2B & SFT & \texttt{osunlp/\allowbreak{}UGround-\allowbreak{}V1-\allowbreak{}2B} \\
VLM & Qwen2-VL-2B & SFT & \texttt{Minthy/\allowbreak{}ToriiGate-\allowbreak{}v0.\allowbreak{}4-\allowbreak{}2B} \\
VLM & Qwen2-VL-2B & SFT & \texttt{weihongliang/\allowbreak{}RC-\allowbreak{}Qwen2VL-\allowbreak{}2b} \\
VLM & SmolVLM-500M-Instruct & base & \texttt{HuggingFaceTB/\allowbreak{}SmolVLM-\allowbreak{}500M-\allowbreak{}Instruct} \\
VLM & SmolVLM-500M-Instruct & SFT & \texttt{YuankaiLuo/\allowbreak{}SimVLA-\allowbreak{}LIBERO} \\
VLM & SmolVLM-500M-Instruct & SFT & \texttt{racineai/\allowbreak{}Flantier-\allowbreak{}SmolVLM-\allowbreak{}500M-\allowbreak{}dse} \\
VLM & SmolVLM-500M-Instruct & SFT & \texttt{BIOMEDICA/\allowbreak{}BMC-\allowbreak{}smolvlm1-\allowbreak{}500M} \\
VLM & SmolVLM-500M-Instruct & SFT & \texttt{TESS-\allowbreak{}Computer/\allowbreak{}smolvlm-\allowbreak{}atari-\allowbreak{}vla} \\
Diffusion & Lumina-Image-2.0 & base & \texttt{Alpha-\allowbreak{}VLLM/\allowbreak{}Lumina-\allowbreak{}Image-\allowbreak{}2.\allowbreak{}0} \\
Diffusion & Lumina-Image-2.0 & SFT & \texttt{duongve/\allowbreak{}NetaYume-\allowbreak{}Lumina-\allowbreak{}Image-\allowbreak{}2.\allowbreak{}0} \\
Diffusion & Lumina-Image-2.0 & SFT & \texttt{neta-\allowbreak{}art/\allowbreak{}Neta-\allowbreak{}Lumina} \\
Diffusion & Lumina-Image-2.0 & SFT & \texttt{X-\allowbreak{}ART/\allowbreak{}LeX-\allowbreak{}Lumina} \\
Diffusion & Lumina-Image-2.0 & SFT & \texttt{OnomaAIResearch/\allowbreak{}Illustrious-\allowbreak{}Lumina-\allowbreak{}v0.\allowbreak{}03} \\
Diffusion & Playground v2.5 & base & \texttt{playgroundai/\allowbreak{}playground-\allowbreak{}v2.\allowbreak{}5-\allowbreak{}1024px-\allowbreak{}aesthetic} \\
Diffusion & Playground v2.5 & SFT & \texttt{waadarsh/\allowbreak{}magnite-\allowbreak{}lora} \\
Diffusion & Playground v2.5 & SFT & \texttt{Jayw1/\allowbreak{}xubeihong\_\allowbreak{}lora\_\allowbreak{}new} \\
Diffusion & Playground v2.5 & SFT & \texttt{Jayw1/\allowbreak{}lora\_\allowbreak{}baimiao} \\
Diffusion & Playground v2.5 & SFT & \texttt{Jayw1/\allowbreak{}lora\_\allowbreak{}xubeihong} \\
Diffusion & Diffusion 3.5 Medium & base & \texttt{stabilityai/\allowbreak{}stable-\allowbreak{}diffusion-\allowbreak{}3.\allowbreak{}5-\allowbreak{}medium} \\
Diffusion & Diffusion 3.5 Medium & SFT & \texttt{ericbill21/\allowbreak{}focus\_\allowbreak{}sd35} \\
Diffusion & Diffusion 3.5 Medium & SFT & \texttt{yandex/\allowbreak{}stable-\allowbreak{}diffusion-\allowbreak{}3.\allowbreak{}5-\allowbreak{}medium-\allowbreak{}alchemist} \\
Diffusion & Diffusion 3.5 Medium & SFT & \texttt{suzushi/\allowbreak{}miso-\allowbreak{}diffusion-\allowbreak{}2.\allowbreak{}1} \\
Diffusion & Diffusion 3.5 Medium & SFT & \texttt{tensorart/\allowbreak{}bokeh\_\allowbreak{}3.\allowbreak{}5\_\allowbreak{}medium} \\
Diffusion & Diffusion v1-4 & base & \texttt{CompVis/\allowbreak{}stable-\allowbreak{}diffusion-\allowbreak{}v1-\allowbreak{}4} \\
Diffusion & Diffusion v1-4 & SFT & \texttt{johnrobinsn/\allowbreak{}sd-\allowbreak{}model-\allowbreak{}gameNgen} \\
Diffusion & Diffusion v1-4 & SFT & \texttt{dixisouls/\allowbreak{}anime-\allowbreak{}diffusion} \\
Diffusion & Diffusion v1-4 & SFT & \texttt{theproparadox/\allowbreak{}stablediffusion\_\allowbreak{}musiccaps} \\
Diffusion & Diffusion v1-4 & SFT & \texttt{Flaaaande/\allowbreak{}sd-\allowbreak{}model-\allowbreak{}mario} \\
Diffusion & Diffusion XL Base 1.0 & base & \texttt{stabilityai/\allowbreak{}stable-\allowbreak{}diffusion-\allowbreak{}xl-\allowbreak{}base-\allowbreak{}1.\allowbreak{}0} \\
Diffusion & Diffusion XL Base 1.0 & SFT & \texttt{RunDiffusion/\allowbreak{}Juggernaut-\allowbreak{}XL-\allowbreak{}v9} \\
Diffusion & Diffusion XL Base 1.0 & SFT & \texttt{RunDiffusion/\allowbreak{}Juggernaut-\allowbreak{}X-\allowbreak{}v10} \\
Diffusion & Diffusion XL Base 1.0 & SFT & \texttt{diffusers/\allowbreak{}stable-\allowbreak{}diffusion-\allowbreak{}xl-\allowbreak{}1.\allowbreak{}0-\allowbreak{}inpainting-\allowbreak{}0.\allowbreak{}1} \\
Diffusion & Diffusion XL Base 1.0 & SFT & \texttt{cagliostrolab/\allowbreak{}animagine-\allowbreak{}xl-\allowbreak{}4.\allowbreak{}0} \\
\end{longtable}
\end{center}

\FloatBarrier

\Cref{tab:vlm_diffusion_registry} lists the public HuggingFace repositories used for the VLM and Diffusion checks.

\FloatBarrier
\begin{table}[H]
\centering
\scriptsize
\caption{\textbf{Per-family VLM and Diffusion composition.} The table lists each cross-domain family, its domain, parameter scale, and the number of SFT descendants used in the VLM and Diffusion checks.}
\label{tab:vlm_diffusion_per_family}
\setlength{\tabcolsep}{3pt}
\begin{tabular}{l l c c}
\toprule
Domain & Family & Params & \#SFT \\
\midrule
VLM & \mbox{PaliGemma2-3B-pt-224} & 3B & 4 \\
VLM & \mbox{Qwen2.5-VL-3B-Instruct} & 3B & 4 \\
VLM & \mbox{Qwen2-VL-2B} & 2B & 4 \\
VLM & \mbox{SmolVLM-500M-Instruct} & 500M & 4 \\
Diffusion & \mbox{Lumina-Image-2.0} & 2B & 4 \\
Diffusion & \mbox{Playground v2.5} & 3.5B & 4 \\
Diffusion & \mbox{Diffusion 3.5 Medium} & 2.5B & 4 \\
Diffusion & \mbox{Diffusion v1-4} & 860M & 4 \\
Diffusion & \mbox{Diffusion XL Base 1.0} & 3.5B & 4 \\
\midrule
\textbf{Total} & n/a & n/a & \textbf{36} \\
\bottomrule
\end{tabular}
\end{table}

\FloatBarrier

\Cref{tab:vlm_diffusion_per_family} reports the cross-domain check set: four VLM families and five Diffusion families. Each family has one base checkpoint and four fine-tuned descendants, contributing $36$ fine-tuned descendants in total. These checkpoints are used for the VLM and Diffusion checks and are kept separate from the LLM headline counts.

\paragraph{Fine-tune diversity.}
The post-training children span a wide range of objectives. The LLM SFT set includes code generation, multilingual translation, mathematical reasoning, tool-use tuning, instruction following, chat tuning, text-to-SQL, medicine, legal drafting, creative writing, and safety-oriented writing. The LLM RLHF and alignment set includes DPO, GRPO, PPO, ORPO, and dataset-specific preference tuning. The VLM checks include OCR, retrieval, medical reasoning, visual grounding, and robotics-oriented fine-tunes. The Diffusion checks include image-style, domain-adaptation, inpainting, and media-specific generation fine-tunes.


\end{document}